\documentclass[10pt,journal,compsoc]{IEEEtran}

\ifCLASSOPTIONcompsoc
  \usepackage[nocompress]{cite}
\else
  \usepackage{cite}
\fi

\usepackage{amsmath,amsfonts}
\usepackage{algorithmic}
\usepackage{algorithm}
\usepackage{array}
\usepackage[caption=false,font=normalsize,labelfont=sf,textfont=sf]{subfig}
\usepackage{textcomp}
\usepackage{stfloats}
\usepackage{url}
\usepackage{verbatim}
\usepackage{graphicx}
\usepackage{bm}
\usepackage{multirow, colortbl, xcolor, booktabs}
\usepackage{makecell}
\usepackage{soul}

\usepackage{tablefootnote}
\usepackage{threeparttable}

\usepackage{bbm}

\definecolor{airforceblue}{rgb}{0.36, 0.54, 0.66}
\definecolor{skyblue}{rgb}{0.53, 0.81, 0.92}
\definecolor{frenchblue}{rgb}{0.0, 0.45, 0.73}

\definecolor{americanrose}{rgb}{1.0, 0.01, 0.24}
\definecolor{myred}{rgb}{0.753, 0.314, 0.275}
\definecolor{myblue}{rgb}{0.0, 0.24, 0.95}
\definecolor{tbl_gray}{gray}{0.85}
\definecolor{greenframe}{rgb}{0.0, 0.784, 0.0}
\definecolor{bluedots}{rgb}{0.0, 1.0, 1.0}
\newcommand{\RNum}[1]{\uppercase\expandafter{\romannumeral #1\relax}}

\newcommand\MYhyperrefoptions{bookmarks=true,bookmarksnumbered=true,
pdfpagemode={UseOutlines},plainpages=false,pdfpagelabels=true,
colorlinks=true,linkcolor={americanrose},citecolor={myblue},urlcolor={myblue}}
\usepackage[\MYhyperrefoptions,pdftex]{hyperref}

\usepackage[capitalize]{cleveref}

\begin{document}

\title{Segment Anything in 3D with Radiance Fields}
%
\author{Jiazhong Cen, Jiemin Fang, Zanwei Zhou, Chen Yang, \\Lingxi Xie, Xiaopeng Zhang, Wei Shen, and Qi Tian,~\IEEEmembership{Fellow,~IEEE}
\IEEEcompsocitemizethanks{\IEEEcompsocthanksitem J. Cen, Z. Zhou, C. Yang, and W. Shen are with MoE Key Lab of Artificial Intelligence, AI Institute, Shanghai Jiao Tong University, Shanghai, 200240, China. (Corresponding author: W. Shen.) \protect\\
\IEEEcompsocthanksitem J. Fang, L. Xie, X. Zhang and Q. Tian are with Huawei Inc., China. (Project lead: J. Fang.)}%
}

\markboth{A SUBMISSION TO IEEE TRANSACTION ON PATTERN ANALYSIS AND MACHINE INTELLIGENCE}%
{Shell \MakeLowercase{\textit{et al.}}: Bare Demo of IEEEtran.cls for Computer Society Journals}
%
\IEEEtitleabstractindextext{
\begin{abstract}
The Segment Anything Model (SAM) emerges as a powerful vision foundation model to generate high-quality 2D segmentation results. This paper aims to generalize SAM to segment 3D objects. Rather than replicating the data acquisition and annotation procedure which is costly in 3D, we design an efficient solution, leveraging the radiance field as a cheap and off-the-shelf prior that connects multi-view 2D images to the 3D space. We refer to the proposed solution as \textbf{SA3D}, short for Segment Anything in 3D. With SA3D, the user is only required to provide a 2D segmentation prompt (\textit{e.g.}, rough points) for the target object in a \textbf{single view}, which is used to generate its corresponding 2D mask with SAM. Next, SA3D alternately performs \textbf{mask inverse rendering} and \textbf{cross-view self-prompting} across various views to iteratively refine the 3D mask of the target object. For one view, mask inverse rendering projects the 2D mask obtained by SAM into the 3D space with guidance of the density distribution learned by the radiance field for 3D mask refinement; Then, cross-view self-prompting extracts reliable prompts automatically as the input to SAM from the rendered 2D mask of the inaccurate 3D mask for a new view. We show in experiments that SA3D adapts to various scenes and achieves 3D segmentation within seconds.
Our research reveals a potential methodology to lift the ability of a 2D segmentation model to 3D. Our code is available at \url{https://github.com/Jumpat/SegmentAnythingin3D}.
\end{abstract}


}

\maketitle

\IEEEdisplaynontitleabstractindextext

\IEEEpeerreviewmaketitle

\section{Introduction}
\IEEEPARstart{T}{he} computer vision community has been pursuing a vision foundation model that can perform basic tasks (\textit{e.g.}, segmentation) in any scenario. In these studies, the Segment Anything Model (SAM)~\cite{sam} stands out as a representative work, due to its ability to segment anything in 2D images. However, how to extend the ability of SAM to 3D scenes remains mostly uncovered. One may choose to replicate the pipeline of SAM to collect and semi-automatically annotate a large set of 3D scenes. Yet obtaining and annotating 3D data densely is much more complex than its 2D counterpart, making this data-driven approach impractical.

We realize that an alternative and efficient solution lies in equipping the 2D foundation model (\textit{i.e.}, SAM) with 3D perception via a 3D representation model. In other words, there is no need to establish a 3D foundation model from scratch. We draw inspirations from radiance fields~\cite{NeRF,InstantNGP,dvgo,tensorf,mipnerf360,3dgs}, a set of novel 3D representations. Radiance fields act as a bridge to connect 2D multi-view images to the 3D world with the differentiable rendering technology. In this paper, we propose the integration of SAM with radiance fields to facilitate 3D segmentation.

\begin{figure}[!t]
    \centering
    \includegraphics[width=\linewidth]{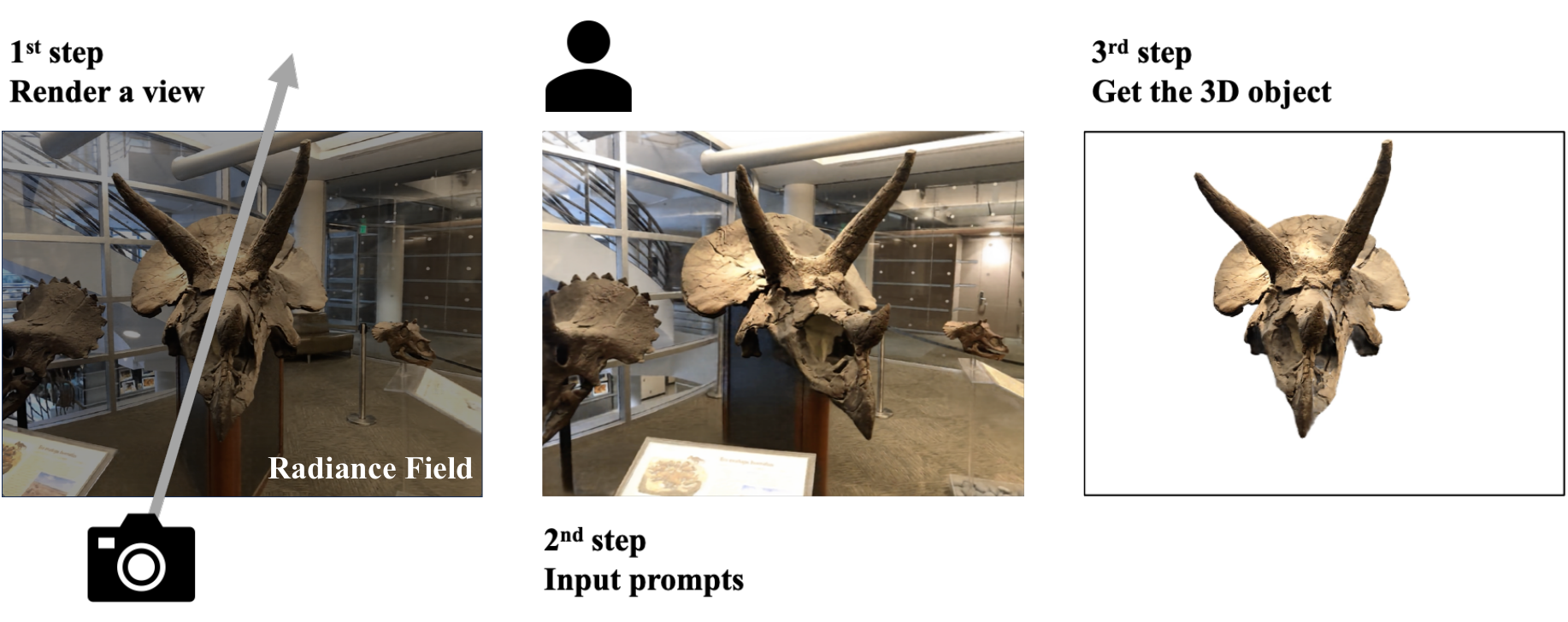}
    \caption{Given a pre-trained radiance field, SA3D takes prompts from one single rendered view as input and outputs the 3D segmentation result for the specific target.}
    \label{fig:teaser}
\end{figure}

As shown in Figure~\ref{fig:teaser}, our solution is named Segment Anything in 3D (\textbf{SA3D}). Given any radiance field trained on a set of 2D images, SA3D takes prompts (\textit{e.g.}, click points on the object) in a single view as input, which is used to generate a 2D mask in this view with SAM. 
Next, SA3D alternately performs two steps across various views to iteratively refine the 3D mask of the object. 
In each round, the first step is \textbf{mask inverse rendering}, in which the previous 2D segmentation mask obtained by SAM is projected into the 3D space via density-guided inverse rendering for 3D mask refinement. 
The second step is \textbf{cross-view self-prompting}, which uses the radiance field to render the 2D segmentation mask (which may be inaccurate) based on the 3D mask from another view. Then a few point prompts are automatically generated from the rendered mask and used to prompt SAM to produce a more complete and accurate 2D mask.
The above procedure is executed iteratively until all necessary views have been sampled.
{With this streamlined pipeline, SA3D can achieve 3D segmentation within minutes. Noticing that the segmentation process of SAM unfolds into two phases, \textit{i.e.} a time-consuming feature extraction phase and a quick mask decoding phase, we further propose to pre-cache the features extracted by the SAM encoder for essential views. This simple modification notably accelerates the inference speed of SA3D. As a result, when integrating with an efficient representation like 3D-GS~\cite{3dgs}, SA3D can segment a 3D object from a scene in high quality within 2 seconds at the fastest.
This acceleration benefits the wider application of SA3D.
}

{We evaluate the performance of SA3D with various segmentation settings (\textit{e.g.}, object and part-level) across multiple datasets~\cite{replica, nvos, spinnerf, lerf, tanks} to demonstrate its effectiveness and efficiency. Without re-training/re-designing SAM or the radiance field, SA3D easily and efficiently adapts to a variety of scenarios, including forward-facing and $360^\circ$ scenes.
SA3D offers an efficient tool for 3D segmentation in radiance fields and reveals a generic methodology to lift 2D segmentation models to the 3D space.
Through comprehensive experiments, we demonstrate the compatibility of SA3D with different radiance fields and provide deeper discussions on its working mechanism.
}


{
The preliminary version of this paper appeared as a conference paper~\cite{sa3d}. We make extensions in this journal version as follows: i) We adapt SA3D for more representative radiance fields, especially 3D-GS~\cite{3dgs}, showing its adaptability and achieving a notable promotion in segmentation speed; ii) We introduce a feature cache mechanism that further enhances segmentation speed by over 6.9 times\footnote{Concrete acceleration depends on specific scenes. See Table~\ref{tab:concrete_time} for more details.} by addressing the speed bottleneck imposed by the SAM forward pass; iii) We provide an expanded set of experimental results, including comprehensive ablation studies that deeply analyze the segmentation performance of SA3D when applied to different radiance fields. These experiments aim to provide deeper insights into the working mechanism of SA3D.
}

{The remaining part of the paper is organized as follows. In Section~\ref{sec:related} we introduce related works including radiance fields, existing 2D / 3D segmentation methods and segmentation in radiance fields. Then we review preliminaries of our method in Section~\ref{sec:pre}. In Section~\ref{sec:meth}, we first introduce the overall pipeline of our proposed SA3D and then explain each components of it in detail. In Section~\ref{sec:experiments}, we evaluate our method quantitatively and qualitatively on various datasets to demonstrate its effectiveness. This section also includes extensive experiments to analyze each component of SA3D in depth. We further provide more discussions about the working mechanism in Section~\ref{sec:dis}. We finally draw a conclude in Section~\ref{sec:conclusion}.}



\section{Related Work}
\label{sec:related}

\subsection{Radiance Fields} {
NeRF (Neural Radiance Field)~\cite{NeRF} represents 3D scenes as continuous functions parameterized by MLPs that map from a 3D coordinate to properties of the scene at the corresponding location. It leverages the differentiable volume rendering technique to translate a 3D scene's continuous representation into 2D images.
Nevertheless, vanilla-NeRF has unfavorable limitations of lengthy training progress and slow rendering speed. These limitations motivate an explosion of follow-up methods devoted to accelerating optimization and rendering~\cite{dvgo,dvgov2,tensorf,InstantNGP, plenoxel, fastnerf, kilonerf, plenoctrees, donerf, dnerf, liang2022spidr}.
Though some breakthroughs have been achieved, the reliance on low-efficient volume rendering still hinders real-time rendering.
Recently, Kerbl \emph{et al.}~\cite{3dgs} proposed a real-time rendering technique, 3D Gaussian Splatting (3D-GS).
Unlike previous studies, it adopts a set of unstructured 3D Gaussians to represent the 3D scene and utilizes an efficient rasterization algorithm to replace the volume rendering. 3D-GS holds on par quality while reaching real-time rendering speed. Thus, it is regarded as a potential game changer in the area of radiance fields. Based on 3D-GS, there are a lot of studies focusing on enhancing its rendering quality, reducing its storage costs, or lifting it to 4D representation~\cite{3dgs, 3dgssurvey, mipsplatting, 4dgs, yang2023real, 4dgs2, luiten2023dynamic, deblur3dgs, fsgs, dynamic3dgs2, gaussianobject}. In this paper, we analyze the segmentation performance of SA3D across representative radiance fields, aiming to demonstrate the adaptability of our method and provide deeper insights into its working mechanism.}

\subsection{2D Segmentation} Since FCN~\cite{fcn} has been proposed, research on 2D image segmentation has experienced a rapid growth. Various sub-fields of segmentation have been explored deeply by numerous studies~\cite{segnet,maskrcnn,panoptic-seg,deeplab,pspnet,shen2023survey}. With transformers~\cite{transformer,vit} entering the field of segmentation, many new segmentation architectures~\cite{setr,maskformer,mask2former,segmenter,segformer} have been proposed and the whole field of segmentation has been further developed. 
A significant breakthrough in this field is the Segment Anything Model (SAM)~\cite{sam}. As an emerging vision foundation model, SAM aims to unify the 2D segmentation task by introducing a prompt-based segmentation paradigm. {The emergence of SAM triggered a new round of research boom, with numerous studies aiming to refine its capabilities. These enhancements range from efficient fine-tuning techniques~\cite{tunesam1,tunesam2,hqsam} to methods for distillation-based acceleration~\cite{fastsam, mobileSAM}. Furthermore, SAM has been extended into various domains, including but not limited to medical image analysis~\cite{medicalSAM, mazurowski2023segment, wu2023medical, deng2023segment}, concealed object detection~\cite{ConcealedSAM, tang2023can}, image editing~\cite{yu2023inpaint}, remote sensing~\cite{chen2024rsprompter} and 3D bird's eye view (BEV) sensing~\cite{zhang2023sam3d}.}
An analogous model to SAM is SEEM~\cite{seem}, which exhibits impressive open-vocabulary segmentation capabilities.

\subsection{3D Segmentation and Perception} 
Numerous methods have explored 3D segmentation with various types of 3D representations. These representations include RGB-D images~\cite{wang2018depth, wu2022depth, xing2020malleable, chu2018surfconv}, point clouds~\cite{qi2017pointnet, qi2017pointnet++, zhao2021point} and grid space such as voxels~\cite{huang2016point, tang2020searching, liu2019point}, cylinders~\cite{zhou2020cylinder3d} and bird's eye view space~\cite{ye2022lidarmultinet, gosala2022bird}. Although 3D segmentation has been developed for a period of time, compared with 2D segmentation, the scarcity of labeled data and high computational complexity makes it difficult to design a segmentation foundation model similar to SAM. {Recently, to tackle this limitation of data scarcity, many studies~\cite{lerf, langsplat, peng2023openscene, ding2023pla, ha2022semantic, zhang2023clip, liu2023partslip, yang2023regionplc, jatavallabhula2023conceptfusion} propose to inject the perception ability of 2D vision-language models (VLM) like CLIP~\cite{clip} into 3D representations by projecting and fusing multi-view 2D visual features with corresponding 3D features. Then, with fused 3D features these methods can perform language-driven 3D perception task like localization and reasoning. Different from these methods, SA3D specializes in providing fine-grained segmentation based on input prompts, focusing more on detailed visual analysis than scene understanding through language.
}

\subsection{Segmentation in Radiance Fields} 
\label{sec:seg_in_rfs}
Inspired by the success of NeRFs in 3D consistent novel view synthesis, numerous studies have delved into the realm of 3D segmentation within NeRFs. 

Zhi \emph{et al.}~\cite{semantic-nerf} propose Semantic-NeRF, a method that incorporates semantics into appearance and geometry, which demonstrates the potential of NeRFs in label propagation and refinement. NVOS~\cite{nvos} introduces an interactive approach to select 3D objects from NeRFs by training a lightweight multi-layer perception (MLP) using custom-designed 3D features. Other approaches, \textit{e.g.} N3F~\cite{n3f}, DFF~\cite{dff}, LERF~\cite{lerf}, LangSplat~\cite{langsplat}, ISRF~\cite{isrf}, NeRF-SOS~\cite{nerfsos}, Feature3DGS~\cite{feature3dgs} and FMGS~\cite{fmgs} aim to lift 2D visual features to 3D through aligning additional feature fields with multi-view 2D features. In inference stage, segmentation is conducted by feature matching between textual features or 2D visual features with 3D features.
Different from the paradigm of feature alignment, Mirzaei \emph{et al.}~\cite{spinnerf} propose MVSeg as a component of their NeRF inpainting framework SPIn-NeRF. MVSeg propagates a 2D mask across different views with the help of 2D self-supervised models like DINO~\cite{dino} and employs these masks as labels for training a Semantic-NeRF model. However, the mask propagation of MVSeg solely relies on 2D visual feature similarities and does not incorporate 3D location information. This omission makes the obtained masks noisy, leading to bad performance of MVSeg in complex scenes. In comparison, the cross-view self-prompting mechanism employed by SA3D adequately addresses this issue.

{Recently, many studies~\cite{saga, gaussiangrouping, garfield, omniseg3d} propose to automatically generate masks from multi-view images within the training set of radiance field models using SAM. Then, these masks are distilled into 3D feature fields with specifically designed loss functions. Once the 3D feature fields are trained, 3D interactive segmentation or grouping can be performed via 3D feature matching. In these methods, all 3D features are derived from automatically extracted masks, suggesting that subsequent segmentation will fail to capture objects overlooked during the automatic mask extraction phase. The absence of human intervention makes it challenging to ensure the accurate segmentation of all objects of interest. In contrast, SA3D can conduct accurate 3D segmentation that aligns properly with user requirements.}

There are also many instance segmentation and semantic segmentation approaches~\cite{obsurf,uorf,rfp, instance-nerf, dmnerf, panopticnerf, nesf} combined with NeRFs. 
These methods predominantly concentrate on segmenting objects belonging to some predefined classes. In contrast, SA3D is able to segment objects belonging to any classes by harnessing the capacity of a 2D segmentation foundation model.

\section{Preliminaries}
\label{sec:pre}
{In this section, we provide a brief background of fundamentals used in SA3D. We first revisit radiance fields in Section~\ref{sec:pre_rf} and then introduce the Segment Anything Model (SAM) in Section~\ref{sec:pre_sam}.}

\subsection{Radiance Fields}
\label{sec:pre_rf}

\subsubsection{Neural Radiance Fields (NeRFs)}
Given a training dataset $\mathcal{I}$ of multi-view 2D images,
NeRFs~\cite{NeRF} learn a function $f_{\bm{\theta}}:(\mathbf{x}, \mathbf{d}) \rightarrow (\mathbf{c}, \sigma)$, which maps the spatial coordinates $\mathbf{x}\in \mathbb{R}^3$ and the view direction $\mathbf{d}\in \mathbb{S}^2$ of a point into the corresponding color $\mathbf{c}\in\mathbb{R}^3$ and volume density $\sigma\in\mathbb{R}$. $\bm{\theta}$ denotes the learnable parameters of the function $f$. 
Generally, to render an image $\mathbf{I}_{\bm{\theta}}$, each pixel undergoes a ray casting process where a ray $\mathbf{r}(t) = \mathbf{x}_{o} + t\mathbf{d}$ is projected through the camera pose. Here, $\mathbf{x}_o$ is the camera origin, $\mathbf{d}$ is the ray direction, and $t$ denotes the distance of a point along the ray from the origin.
The RGB color $\mathbf{I}_{\bm{\theta}}(\mathbf{r})$ at the location determined by ray $\mathbf{r}$ is obtained via the differentiable volume rendering algorithm:
\begin{equation}
\label{eq:volume_rendering}
    \mathbf{I}_{\bm{\theta}}(\mathbf{r}) = \int_{t_n}^{t_f}\omega(\mathbf{r}(t))\mathbf{c}(\mathbf{r}(t), \mathbf{d}) \rm{d}t,
\end{equation}
where $\omega(\mathbf{r}(t)) = \mathrm{exp}(-\int_{t_n}^{t} \sigma(\mathbf{r}(s))\mathrm{d}s) \cdot \sigma(\mathbf{r}(t))$, and $t_n$ and $t_f$ denote the near and far bounds of the ray, respectively.

\subsubsection{3D Gaussian Splatting (3D-GS)} {As a recent advancement of radiance field, 3D-GS has many different attributes.  
Rather than learning the function $f_{\bm{\theta}}$ for query-based rendering, 3D-GS adopts a set of 3D Gaussians $\mathcal{G}$ to preserve the properties (\emph{e.g.}, color and opacity) required by rendering. The mean of each Gaussian represents its position in the 3D space and the covariance represents the scale. Given a specific camera pose, 3D-GS projects the 3D Gaussians to 2D and then computes the color $\mathbf{C}$ of a ray by blending a set of ordered Gaussians $\mathcal{G}_{\mathbf{r}}$ overlapping the ray $\mathbf{r}$. Let $\mathbf{g}_i$ denote the i-th Gaussian in $\mathcal{G}_{\mathbf{r}}$, this process is formulated as:
\begin{equation}
\label{eq:rasterization}
    \mathbf{I}_{\bm{\theta}}(\mathbf{r})=\sum_{i = 1}^{|\mathcal{G}_{\mathbf{r}}|}\omega_{\mathbf{g}_i}\mathbf{c}_{\mathbf{g}_i},
\end{equation}
where $\omega_{\mathbf{g}_i} = \alpha_{\mathbf{g}_i}\prod_{j=1}^{i-1}(1-\alpha_{\mathbf{g}_j})$ and $\mathbf{c}_{\mathbf{g}_i}$ 
is the color of Gaussian $\mathbf{g}_i$. $\alpha_{\mathbf{g}_i}$ is given by evaluating a 2D Gaussian with covariance $\Sigma$ multiplied with a learned per-Gaussian opacity. 
The introduction of 3D Gaussians and the rasterization algorithm enables 3D-GS to avoid querying empty spaces during rendering. This leads to significant improvements in rendering and training speed.}

\subsection{Segment Anything Model (SAM)}
\label{sec:pre_sam}
SAM~\cite{sam} employs an encoder-decoder architecture. The encoder $S_e$ takes an image $\mathbf{I}$ as input and outputs the corresponding features $\mathbf{f}_{\mathbf{I}}$:
\begin{equation}
   \mathbf{f}_{\mathbf{I}} = S_e (\mathbf{I}). 
\end{equation}
The decoder $S_d$ takes the features and a set of prompts $\mathcal{P}$ as input, and outputs the corresponding 2D segmentation mask $\mathbf{M}_{\texttt{SAM}}$ in the form of a bitmap, \emph{i.e.,}
\begin{equation}
    \mathbf{M}_{\texttt{SAM}} = S_d (\mathbf{f}_{\mathbf{I}}, \mathcal{P}). 
\end{equation}
The prompts $\mathbf{p} \in \mathcal{P}$ can be points, boxes, texts, and masks. In the computational overhead of SAM, the majority is expended on the encoder, and the decoder accounts for only a minimal portion.

\section{Method}
\label{sec:meth}
In this section, we first introduce the overall pipeline of SA3D. Then, we explain the design of each component in SA3D in detail.

\begin{figure*}[!t]
    \centering
    \includegraphics[width=0.96\linewidth]{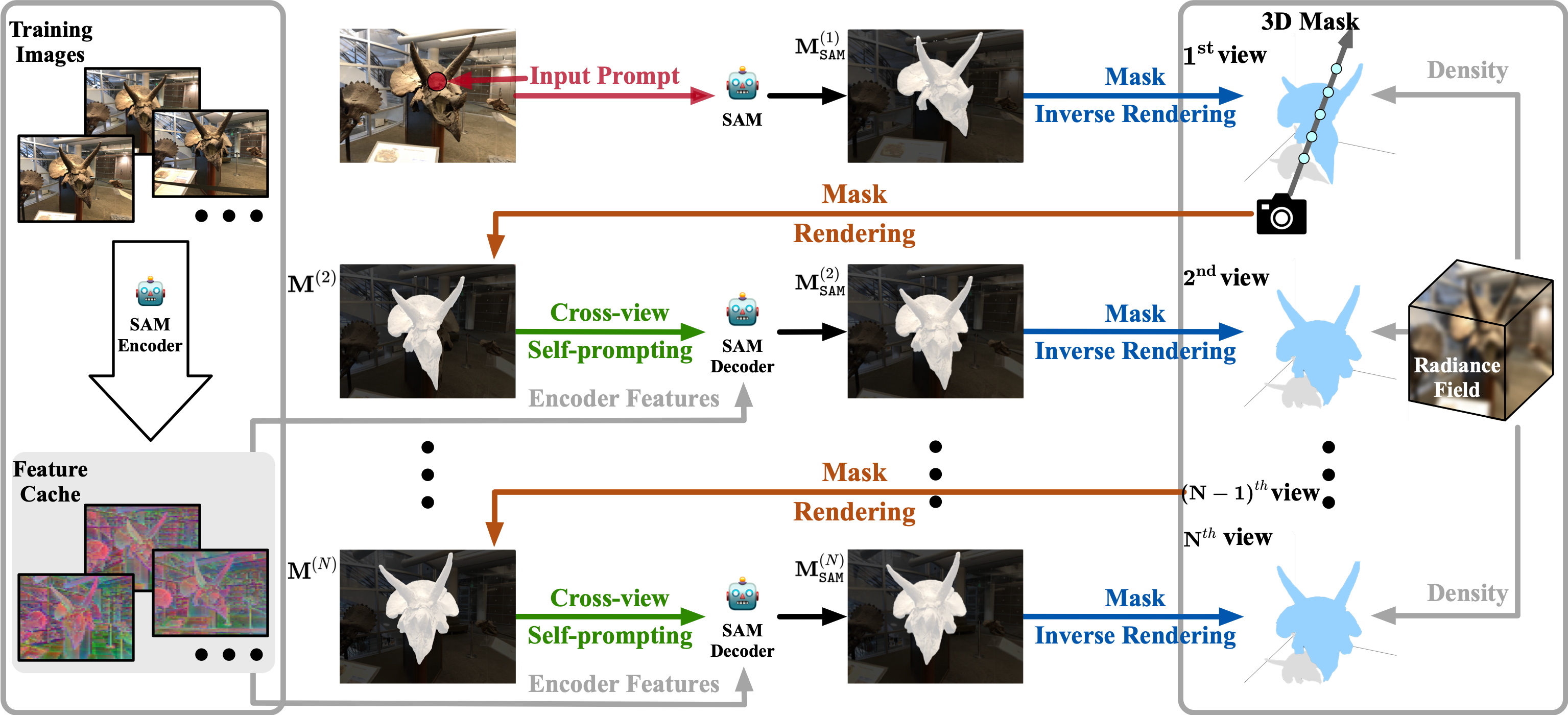}
    \caption{The overall pipeline of SA3D. Given a set of multi-view 2D images and a radiance field trained on it, SA3D first employs the SAM encoder to extract all features from the images and builds a cache. Then SA3D takes prompts in a single view for the target object as input and uses SAM to produce a 2D mask in this view with these prompts. Subsequently, SA3D performs an alternated process of \textbf{mask inverse rendering} and \textbf{cross-view self-prompting} to refine the 3D mask of the target object. Mask inverse rendering is performed to project the 2D mask obtained by the SAM decoder into the 3D space according to the learned density distribution embedded in the radiance field for 3D mask refinement. Cross-view self-prompting is conducted to extract reliable prompts automatically as the input to the SAM decoder from the rendered 2D mask given a novel view. This alternated process is executed iteratively until we get the accurate 3D mask.}
    \label{fig:pipeline}
\end{figure*}

\subsection{Overall Pipeline}
\label{sec:overall}
We assume that we already have a radiance field trained on a set of multi-view images $\mathcal{I}$.
As shown in Figure~\ref{fig:pipeline}, before the segmentation process begins, the SAM encoder $S_e$ is first employed to extract the features of all images in $\mathcal{I}$, which are used to construct a feature cache $\mathcal{F} = \{\mathbf{f}_{\mathbf{I}} | \mathbf{I} \in \mathcal{I}\}$.

At the beginning of segmentation, an image $\mathbf{I}^{(1)}$ from a specified view is rendered with the pre-trained radiance field. A set of prompts (\textit{e.g.}, in this paper, we often use a set of points), $\mathcal{P}^{(1)}$, is introduced and fed into SAM along with the rendered image. The 2D segmentation mask $\mathbf{M}^{(1)}_{\texttt{SAM}}$ of the according view is obtained, which is then projected into the 3D space to form a coarse 3D mask with the proposed \textbf{mask inverse rendering} technique (Section~\ref{sec:optimization}). 
Then a 2D segmentation mask $\mathbf{M}^{(n)}$ from a new view in $\mathcal{I}$ is rendered from the 3D mask. The rendered mask may be inaccurate, especially in the preliminary stage. We propose a \textbf{cross-view self-prompting} method (Section~\ref{sec:prompt}) to extract point prompts $\mathcal{P}^{(n)}$ from the rendered mask. These extracted prompts, combined with the cached features $\mathbf{f}_{\mathbf{I}^{(n)}}$ of the corresponding image $\mathbf{I}^{(n)}$, are then fed into the SAM decoder to generate a refined 2D segmentation mask $\mathbf{M}_{\texttt{SAM}}^{(n)}$. This mask is also projected into the 3D space for refinement. The above procedure is executed iteratively with more views traversed. Meanwhile, the 3D mask becomes more and more accurate. The whole process bridges 2D segmentation results with 3D ones efficiently. Note that no other parameter needs to be optimized except for the 3D mask.

\subsection{3D Mask Representation}
\label{sec:3dmask}
{We design two 3D mask representations to fit the two types of radiance fields, \textit{i.e.} NeRFs and 3D-GS, for better efficiency and effectiveness.}

{For NeRFs, we simply adopt 3D voxel grids $\mathbf{V} \in \mathbb{R}^{L\times W\times H}$ to represent the 3D mask, where each grid vertex stores a zero-initialized soft mask confidence score. Based on these voxel grids, each pixel of the 2D mask from one view is rendered as
\begin{equation}
\label{eq:mask_rendering}
    \mathbf{M}(\mathbf{r}) = \int_{t_n}^{t_f}\omega(\mathbf{r}(t))\mathbf{V}(\mathbf{r}(t)) \rm{d}t,
\end{equation}
where $\mathbf{V}(\mathbf{r}(t))$ denotes the mask confidence score at the location $\mathbf{r}(t)$ obtained from voxel grids $\mathbf{V}$\footnote{$\mathbf{V}(\mathbf{r}(t))$ is computed by trilinearly interpolating vertex values of the 3D mask grids.}.}

{For 3D-GS, we adopt a slightly different design, where a zero-initialized mask confidence score $m_{\mathbf{g}_i}$, is integrated as a new attribute for each 3D Gaussian $\mathbf{g}_i$. 
This is because that in 3D-GS, a 3D Gaussian is a basic unit for rendering. This requires a Gaussian to be wholly assigned to either the segmentation target or the background. Thus, there is no benefit to use additional mask grids for mask representation.
Under this design, the mask rendering process can seamlessly use the differentiable rasterization algorithm as described in Equation~\eqref{eq:rasterization}:
\begin{equation}
\label{eq:mask_rasterization}
    \mathbf{M}(\mathbf{r})=\sum_{i=1}^{|\mathcal{G}_{\mathbf{r}}|}\omega_{\mathbf{g}_i}m_{\mathbf{g}_i}.
\end{equation}
Moreover, this modification brings the following advantages: 1) Reduced GPU memory usage. The number of 3D Gaussians in a scene rarely surpasses 10 million, contrasting sharply with the 32.77 million entries required by a $320^3$ resolution mask grid. Attaching an attribute to each Gaussian is thus significantly more memory-efficient; 2) Lower computation consumption. Storing mask confidence in grids necessitates additional querying during the rendering phase to access each Gaussian's confidence score, thereby increasing the computational overhead.}

\subsection{Mask Inverse Rendering}
\label{sec:optimization}
As shown in Equation~\eqref{eq:volume_rendering} and~\eqref{eq:rasterization}, the color of each pixel in a rendered image $\mathbf{I}$ is determined by a sum of weighted colors along the corresponding ray. The weight $\omega(\mathbf{r}(t))$ (or $\{\omega_{\mathbf{g}_i} | \mathbf{g}_i \in \mathcal{G}_{\mathbf{r}}\}$ for 3D-GS) reveals the object structure within the 3D space, where a high weight indicates the corresponding point close to the object's surface. Mask inverse rendering aims to project the 2D mask of $\mathbf{I}$ into the 3D space to form the 3D mask based on these weights.

Denote $\mathbf{M}_{\texttt{SAM}}(\mathbf{r})$ as the corresponding mask generated by SAM. When $\mathbf{M}_{\texttt{SAM}}(\mathbf{r}) = 1$, the goal of mask inverse rendering is to increase $\mathbf{V}(\mathbf{r}(t))$ (or $\{m_{\mathbf{g}_i} | \mathbf{g}_i \in \mathcal{G}_{\mathbf{r}}\}$ for 3D-GS) with respect to $\omega$.
In practice, this can be optimized using the gradient descent algorithm. For this purpose, we define the mask projection loss as the negative product between $\mathbf{M}_{\texttt{SAM}}(\mathbf{r})$ and $\mathbf{M}(\mathbf{r})$:
\begin{equation}
    \label{eq:loss_p}
    \mathcal{L}_{\texttt{proj}} = -\sum_{\mathbf{r}\in \mathcal{R}(\mathbf{I})}\mathbf{M}_{\texttt{SAM}}(\mathbf{r}) \cdot \mathbf{M}(\mathbf{r}),
\end{equation}
where $\mathcal{R}(\mathbf{I})$ denotes the ray set of the image $\mathbf{I}$.

The mask projection loss is constructed based on the assumption that both the geometry from the radiance field and the segmentation results of SAM are accurate. However, in practice, this is not always the case. We append a negative refinement term to the loss to optimize the 3D mask grids according to multi-view mask consistency:
\begin{equation}
\begin{aligned}
    \mathcal{L}_{\texttt{proj}} = &-\sum_{\mathbf{r}\in \mathcal{R}(\mathbf{I})} \mathbf{M}_{\texttt{SAM}}(\mathbf{r}) \cdot \mathbf{M}(\mathbf{r}) \\ &+ \lambda \sum_{\mathbf{r}\in \mathcal{R}(\mathbf{I})} (1-\mathbf{M}_{\texttt{SAM}}(\mathbf{r})) \cdot \mathbf{M}(\mathbf{r}), 
\end{aligned}
    \label{eq:loss_p2}
\end{equation}
where $\lambda$ is a hyper-parameter to determine the magnitude of the negative term. With this negative refinement term, only if SAM consistently predicts a region as foreground from different views, SA3D marks its corresponding 3D region as foreground.

\subsection{Cross-view Self-prompting}
\label{sec:prompt}

Mask inverse rendering enables projecting 2D masks into the 3D space to form the 3D mask of a target object. To construct an accurate 3D mask, substantial 2D masks from various views need to be projected. SAM can provide high-quality segmentation results given proper prompts. However, manually selecting prompts from every view is time-consuming and impractical. We propose a cross-view self-prompting mechanism to produce prompts for different views automatically.

Specifically, we first render a 2D segmentation mask $\mathbf{M}^{(n)}$ for a new view from the 3D mask according to Equation~\eqref{eq:mask_rendering} (or Equation~\eqref{eq:mask_rasterization} for 3D-GS). This mask is usually inaccurate, especially at the preliminary iteration of SA3D. Then we obtain some point prompts from the rendered mask with a specific strategy. The above process is named cross-view self-prompting. While there are multiple possible solutions for this strategy, we present a feasible one that has been demonstrated to be effective.

\begin{figure}
    \centering
    \includegraphics[width=0.95\linewidth]{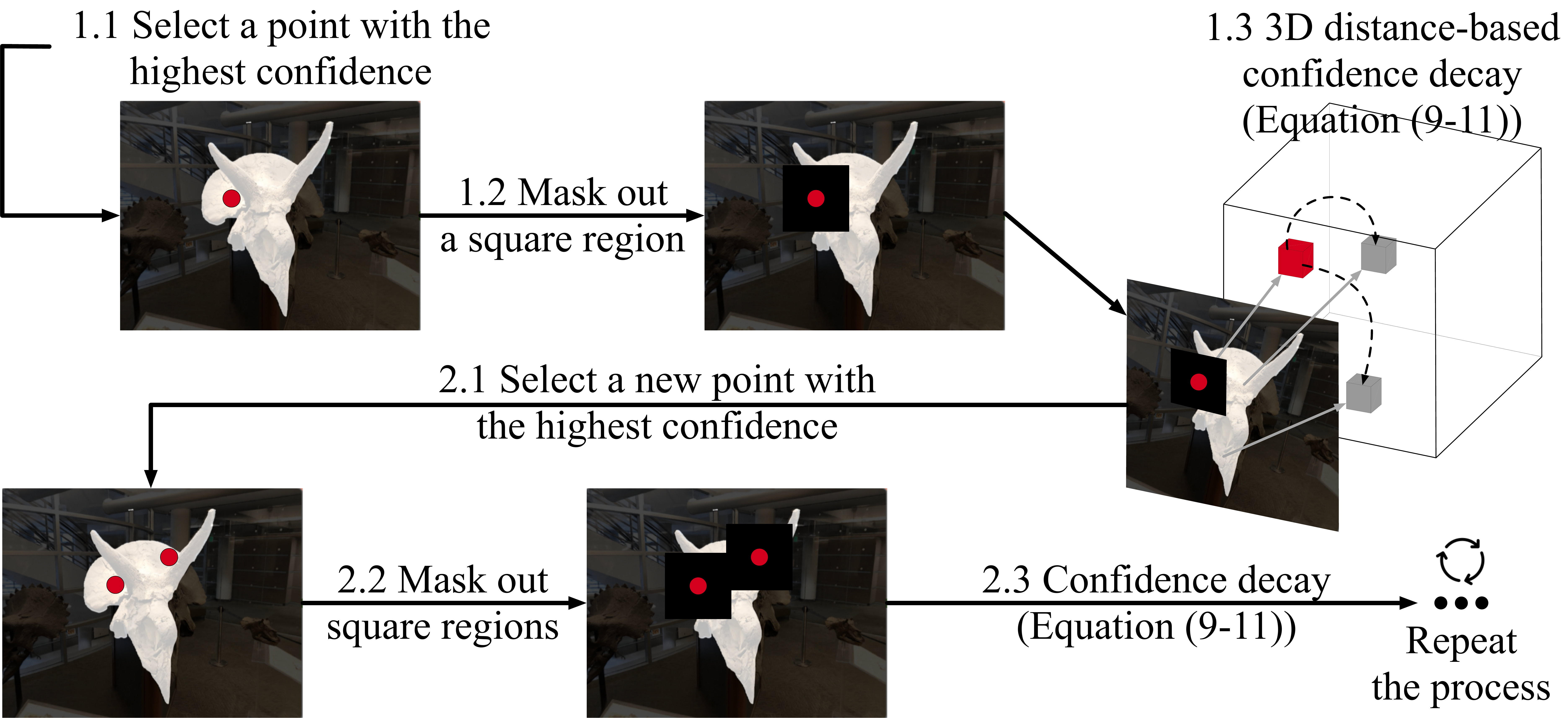}
    \caption{Illustration of the proposed self-prompting strategy. It converts the 2D-rendered mask of a new view into corresponding point prompts, leveraging the 3D geometry prior and 2D-rendered mask confidence.}
    \label{fig:mask_to_prompt}
\end{figure}

\subsubsection{Self-prompting Strategy}
Given an inaccurate 2D rendered mask $\mathbf{M}^{(n)}$, the self-prompting strategy aims to extract a set of prompt points $\mathcal{P}_s$ from it, which can help SAM to generate 2D segmentation result as accurate as possible.
It is important to note that $\mathbf{M}^{(n)}$ is not a typical 2D bitmap, but rather a confidence score map computed using Equation~\eqref{eq:mask_rendering}. We propose a simple yet effective self-prompting strategy using this confidence score and the 3D geometry prior provided by the radiance field, as shown in Figure~\ref{fig:mask_to_prompt}.
Since each image pixel $\mathbf{p}$ corresponds to a ray $\mathbf{r}$ in a rendered view, we use $\mathbf{p}$ for an easier demonstration of the prompt selection strategy on images.

As $\mathcal{P}_s$ is initialized to an empty set, the first prompt point $\mathbf{p}_0$ is selected as the point with the highest mask confidence score: $\mathbf{p}_0 = \arg\max_{\mathbf{p}}\{\mathbf{M}^{(n)}(\mathbf{p}) | \mathbf{p} \in \mathcal{R}(\mathbf{I}^{(n)}) \}$\footnote{$\mathbf{I}^{(n)}$ is the RGB image of the corresponding view. $\mathcal{R}(\mathbf{I}^{(n)})$ denotes rays within $\mathbf{I}^{(n)}$. We use pixel $\mathbf{p}$ rather than ray $\mathbf{r}$ for clarity and brevity since pixels have one-to-one correspondence with rays.}.
To select new prompt points, we first mask out square-shaped regions\footnote{The side length of the region is set as the radius of a circle whose area is equal to the binarized rendered mask $\Bar{\mathbf{M}}^{(n)}$, where $\Bar{\mathbf{M}}^{(n)}(\mathbf{p}) = \mathbbm{1}(\mathbf{M}^{(n)}(\mathbf{p}) > 0)$. $\mathbbm{1}$ denotes the indicator function.} on $\mathbf{M}^{(n)}$ centered with each existing point prompt $\mathbf{\hat{p}} \in \mathcal{P}_s$. 
Considering the depth $z(\mathbf{p})$ can be estimated by the pre-trained NeRF, we transform the 2D pixel $\mathbf{p}$ to the 3D point $\mathcal{G}(\mathbf{p}) = [x(\mathcal{G}(\mathbf{p})), y(\mathcal{G}(\mathbf{p})), z(\mathcal{G}(\mathbf{p})]^\top$:

\begin{equation}
    \begin{pmatrix}
        x(\mathcal{G}(\mathbf{p})) \\
        y(\mathcal{G}(\mathbf{p})) \\
        z(\mathcal{G}(\mathbf{p}))\\
    \end{pmatrix}
         =
         z(\mathbf{p})
         \mathbf{K}^{-1}
    \begin{pmatrix}
        x(\mathbf{p})\\
        y(\mathbf{p})\\
        1\\
    \end{pmatrix}
\end{equation}
where $x(\mathbf{p}), y(\mathbf{p})$ denote the 2D coordinates of $\mathbf{p}$, and $\mathbf{K}$ denotes the camera intrinsics. 
The new prompt point is expected to have a high confidence score while being close to existing prompt points. Considering the two factors, we introduce a decay term to the confidence score.
Let $d(\cdot, \cdot)$ denote the min-max normalized Euclidean distance. For each remaining point $\mathbf{p}$ in $\mathbf{M}^{(n)}$, the decay term is
\begin{equation}
    \label{eq:decay}
    \Delta \mathbf{M}^{(n)}(\mathbf{p}) = \min \{\mathbf{M}^{(n)}(\mathbf{\hat{p}}) \cdot d(\mathcal{G}(\mathbf{p}),\mathcal{G}(\mathbf{\hat{p}}))\ |\ \mathbf{\hat{p}} \in \mathcal{P}_s\}.
\end{equation}
Then a decayed mask confidence score $\Tilde{\mathbf{M}}^{(n)}(\mathbf{p})$ is computed as
\begin{equation}
    \label{eq:decayed_score}
    \Tilde{\mathbf{M}}^{(n)}(\mathbf{p}) = \mathbf{M}^{(n)}(\mathbf{p}) - \Delta \mathbf{M}^{(n)}(\mathbf{p}).
\end{equation}
The remaining point with the highest decayed score, \emph{i.e.}, $\mathbf{p}^\ast=\arg\max_{\mathbf{p}} \{\Tilde{\mathbf{M}}^{(n)}(\mathbf{p}) | \mathbf{p} \in \mathcal{R}(\mathbf{I}^{(n)}) \} $, is added to the prompt set: $\mathcal{P}_s=\mathcal{P}_s\cup\{\mathbf{p}^\ast\}$.
The above selection process is repeated until either the number of prompts $|\mathcal{P}_s|$ reaches a predefined threshold $n_p$ or the maximum value of $\Tilde{\mathbf{M}}^{(n)}(\mathbf{p})$ is smaller than 0. 

\subsubsection{IoU-aware View Rejection} 
When the target object is rendered in heavily occluded views, SAM may produce incorrect segmentation results and degrade the quality of the 3D mask. To avoid such situations, we introduce an additional view rejection mechanism based on the intersection-over-union (IoU) between the rendered mask $\mathbf{M}^{(n)}$ and the SAM prediction $\mathbf{M}^{(n)}_{\texttt{SAM}}$. If the IoU falls below a predefined threshold $\tau$, it indicates a poor overlap between the two masks. The prediction from SAM is rejected, and the mask inverse rendering step is skipped in this iteration.

\begin{figure}
    \centering
    \includegraphics[width=\linewidth]{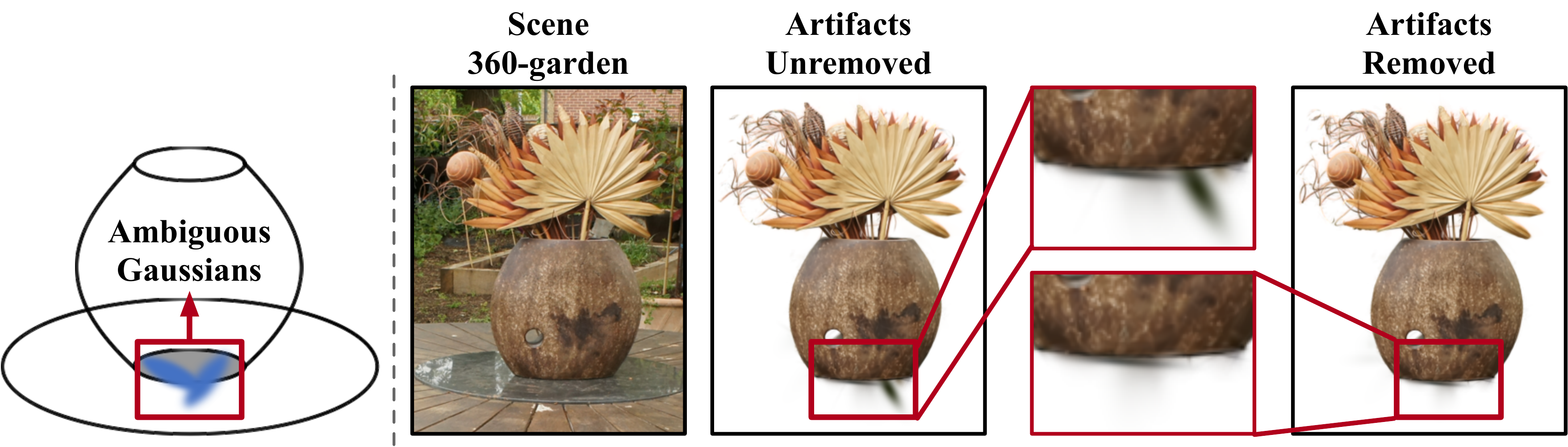}
    \caption{Illustration of ambiguous Gaussians at the interface between two objects. The proposed ambiguous Gaussians removal method can alleviate such phenomenon effectively.}
    \label{fig:artifacts}
\end{figure}

\subsection{Eliminating Ambiguous Gaussians in 3D-GS}
\label{sec:ambi}
{
In 3D-GS, an intuitive way to retrieve the segmentation target is to discard 3D Gaussians with a mask confidence score below a predetermined threshold (0 in default). However, since the 3D Gaussians are trained to fix multi-view RGB images without considering the concept of objects, there are multiple ambiguous Gaussians at the interface between two objects. As shown in Figure~\ref{fig:artifacts}, after the background is removed, these Gaussians appear and impair the geometry correctness of the segmentation target. We provide an optional removal strategy to eliminate these ambiguous Gaussians.}

{
During the segmentation process, we retain all 2D masks extracted by SAM and their corresponding Intersection over Union (IoU) scores used for the IoU-aware view rejection mechanism. After completing segmentation, we remove 3D Gaussians classified as background and reapply the preserved masks for another round of inverse rendering. Views with IoU scores below the threshold $\tau$ remain to be discarded. Since in this step, the occlusions are removed, the negative refinement term in Equation~\eqref{eq:loss_p2} can detect and eliminate most of ambiguous Gaussians.}

{Note that this operation is only adopted in qualitative experiments. The ambiguous Gaussians do not affect the quantitative results before the background is removed.}

\section{Experiments}
\label{sec:experiments}
In this section, we integrate SA3D with different representative radiance fields including vanilla-NeRF~\cite{NeRF}, TensoRF~\cite{tensorf}, and 3D-GS~\cite{3dgs} to form three variants, \emph{i.e.}, SA3D-NeRF, SA3D-TensoRF and SA3D-GS. Vanilla-NeRF employs an MLP to model the 3D scene and is a kind of pure implicit representation. TensoRF employs Vector-Matrix (VM) decomposition and a lightweight MLP decoder to construct the 3D scene, which is a hybrid representation. 3D-GS employs the explicit 3D Gaussians without any neural networks to represent the 3D scene. Experiments demonstrate that SA3D can be adapted to various kinds of radiance fields seamlessly. We quantitatively and qualitatively evaluate the segmentation performance of SA3D on various datasets, further demonstrating the versatility of SA3D, which can conduct instance segmentation, part segmentation, and text-prompted segmentation \textit{etc}.

\subsection{Datasets}
For quantitative experiments, we use the Neural Volumetric Object Selection (NVOS)~\cite{nvos}, SPIn-NeRF~\cite{spinnerf}, and Replica~\cite{replica} datasets. 
The NVOS~\cite{nvos} dataset is based on the LLFF dataset~\cite{llff}, which includes several forward-facing scenes. For each scene, NVOS provides a reference view with scribbles and a target view with 2D segmentation masks annotated.
Similar to NVOS, SPIn-NeRF~\cite{spinnerf} annotates some data manually to evaluate interactive 3D segmentation performance. These annotations are based on some widely-used NeRF datasets~\cite{llff, NeRF, nerf-sup, tanks, plenoxel}.
The Replica~\cite{replica} dataset provides high-quality reconstruction ground truths of various indoor scenes, including clean dense geometry, high-resolution and high-dynamic-range textures, glass and mirror surface information, semantic classes, planar segmentation, and instance segmentation masks. 
For qualitative analysis, we use the LLFF~\cite{llff} dataset and the MIP-360 dataset~\cite{mipnerf360}. SA3D is further applied to the LERF~\cite{lerf} dataset, which contains more realistic and challenging scenes.

\subsection{Implementation Details}
\label{sec:imp_details}
{
We implement SA3D using PyTorch~\cite{pytorch}. The SA3D model is built and trained on a single Nvidia Geforce RTX3090 GPU. Vanilla-NeRF~\cite{NeRF} is pre-trained with 200,000 iterations for all scenes. TensoRF~\cite{tensorf} is pre-trained with 20,000 iterations for LLFF~\cite{llff} and MIP-360~\cite{mipnerf360} and 40,000 iterations for other datasets. For all scenes we choose the VM-48 variant of TensoRF. 3D-GS~\cite{3dgs} is pre-trained with 30,000 iterations for all scenes. During the training of the 3D-GS, we reuse all hyper-parameters from its official implementation.}

{
During segmentation, the resolution of mask grids for NeRFs is set to $320^3$. For vanilla-NeRF, due to the limitations of the GPU memory, the mask of a view is rendered in chunks (32768 rays in each chunk). The learning rates of the mask grids and the mask confidence scores for 3D-GS are set to 1. In default, the coefficient $\lambda$ for the negative refinement term is set to 0.15, the number of prompts $n_p$ is set to 3 and the threshold $\tau$ for the IoU-aware rejection mechanism is set to $0.5$.}

\subsection{Quantitative Results}
\label{sec:quantitative}
\begin{table}
  \caption{Quantitative results on NVOS. `FC' denotes the feature cache. The time cost of compared methods is estimated according to the description in their papers. }
  \label{tab:nvos}
  \centering
  \begin{threeparttable}
  \setlength{\tabcolsep}{1.5mm}{
  \begin{tabular}{lccc}
    \toprule
    Method     & mIoU (\%)& mAcc (\%) & Time (s)\\
    \midrule
    Graph-cut (3D)~\cite{GrabCut,nvos} & 39.4 & 73.6 & -\\
    NVOS~\cite{nvos} & 70.1  & 92.0 & -\\
    ISRF~\cite{isrf} & 83.8 & 96.4 & 150\tnote{*} + 2.3\\
    OmniSeg3D~\cite{omniseg3d} & 91.7 & 98.4 & 900\tnote{*} + 0.1\\
    SA3D-TensoRF (w/o FC)~\cite{sa3d} & 90.3 & 98.2 & 45 \\
    SA3D-TensoRF (w/ FC) & 90.1 & 98.2 & 30\tnote{*} + 15\\
    SA3D-NeRF (w/ FC) & 88.5 & 97.8 & 30\tnote{*} + 200\\
    SA3D-GS (w/ FC) & \textbf{92.2} & \textbf{98.5} & 30\tnote{*} + 2\\
    \bottomrule
  \end{tabular}}
  \begin{tablenotes}
    \item[*] The additional training time cost by ISRF and OmniSeg3D to train feature fields. SA3D does not involve additional training, so we report the time cost of feature caching instead.
    \end{tablenotes}
    \end{threeparttable}
\end{table}

\subsubsection{The NVOS Dataset}
For fair comparisons, we follow the experimental setting of the original NVOS~\cite{nvos}. We first scribble on the reference view (provided by the NVOS dataset) to conduct 3D segmentation, and then render the 3D segmentation result on the target view and evaluate the IoU and pixel-wise accuracy with the provided ground truth. Note that the scribbles are preprocessed to meet the requirements of SAM. As shown in Table~\ref{tab:nvos}, SA3D outperforms previous approaches by large margins, \textit{i.e.}, +0.5 mIoU over the OmniSeg3D, +6.5 mIoU over the ISRF and +20.2 mIoU over the NVOS. The results of SA3D-TensoRF without feature cache are inherited from~\cite{sa3d}. Introducing the feature cache does not affect the segmentation accuracy. The slight number discrepancies in the result tables are due to refinements in our implementation of cross-view self-prompting.

\begin{table*}[htbp]

  \caption{Quantitative results on the SPIn-NeRF dataset. `FC' denotes the feature cache. `TRF' denotes `TensoRF'.}
  \label{tab:spin-nerf}
  \centering
  \setlength{\tabcolsep}{1.45mm}{
  \begin{tabular}{lcc|cc|cc|cc|cc|cc}
    \toprule
    \multirow{2}{*}{Scenes} &  \multicolumn{2}{c}{Single view~\cite{sa3d}} & \multicolumn{2}{c}{MVSeg~\cite{spinnerf}} & \multicolumn{2}{c}{SA3D-TRF (w/o FC)~\cite{sa3d}} & \multicolumn{2}{c}{SA3D-TRF (w/ FC)} & \multicolumn{2}{c}{SA3D-NeRF (w/ FC)}& \multicolumn{2}{c}{SA3D-GS (w/ FC)}\\
    \cmidrule{2-13}
            & IoU (\%)     & Acc (\%) & IoU (\%)     & Acc (\%) & IoU (\%)     & Acc (\%)  & IoU (\%)     & Acc (\%)  & IoU (\%)     & Acc (\%)  & IoU (\%)     & Acc (\%) \\
    \midrule
    Orchids  & 79.4 & 96.0 & \textbf{92.7}  & 98.8& 83.6 & 96.9 & 87.9 & 97.8& 70.4 & 93.3& 84.7 & 97.2\\
    Leaves   & 78.7 & 98.6 & 94.9 & 99.7 &  97.2& 99.9 & \textbf{97.5} & 99.9& 96.4 & 99.8& 97.2 & 99.8\\
    Fern     & 95.2 & 99.3 & 94.3 & 99.2 &  97.1& 99.6 & \textbf{97.3} & 99.6& 93.1 & 99.0& 96.7 & 99.5\\
    Room     & 73.4 & 96.5 & \textbf{95.6} & 99.4 &  88.2& 98.3 & 90.4 & 98.6& 87.5 & 98.3& 93.7 & 99.2\\
    Horns    & 85.3 & 97.1 & 92.8 & 98.7 &   94.5&99.0 & 95.4 & 99.2& \textbf{95.6} & 99.2& 95.3 & 99.2\\
    Fortress & 94.1 & 99.1 & 97.7 & 99.7 &   98.3&99.8 & \textbf{98.4} & 99.8& 97.1 & 99.6& 98.1 & 99.7\\
    \midrule
    Fork     & 69.4 & 98.5 & 87.9 & 99.5 &   89.4 & 99.6 & \textbf{89.8} & 99.6 & 86.4 & 99.4 & 87.9 & 99.5\\
    Pinecone & 57.0 & 92.5 & \textbf{93.4} & 99.2 &   92.9&99.1 & 93.1 & 99.1& 90.9 & 98.9& 91.6 & 99.0\\
    Truck    & 37.9 & 77.9 & 85.2 & 95.1 &  90.8&96.7 & \textbf{96.1} & 98.7& 80.2 & 92.9 & 94.8 & 98.2\\
    Lego     & 76.0 & 99.1 & 74.9 & 99.2 &   \textbf{92.2}&99.8 & 90.9 & 99.7& 86.8 & 99.6& 92.0 & 99.7\\
    \midrule
    mean & 74.6 & 95.5& 90.9& 98.9& 92.4& 98.9 & \textbf{93.7} & 99.2& 88.4 & 98.0& 93.2 & 99.1\\
    \bottomrule
  \end{tabular}}
\end{table*}

\subsubsection{The SPIn-NeRF Dataset} We follow SPIn-NeRF~\cite{spinnerf} to conduct label propagation for evaluation. Given a specific reference view of a target object, the 2D ground-truth mask of this view is available. The prompt input operation is omitted, while the 2D ground-truth mask of the target object from the reference view is directly used for the initialization of the 3D mask grids. This is reasonable since in most situations users can refine their input prompts to help SAM generate a 2D mask as accurately as possible from the reference view. Once the 3D mask grids are initialized, the subsequent steps are exactly the same as described in Section~\ref{sec:meth}. 
With the 3D mask finalized, the 2D masks in other views are rendered to calculate the IoU with the 2D ground-truth masks.
Results can be found in Table~\ref{tab:spin-nerf}. SA3D is demonstrated to be superior in both forward-facing and 360$^{\circ}$ scenes. {The performance of SA3D-NeRF is marginally inferior to that of its counterparts utilizing different radiance fields, as shown in both Table~\ref{tab:nvos} and Table~\ref{tab:spin-nerf}. This discrepancy is attributed to the sub-optimal 3D geometry learned by the Vanilla-NeRF. Specifically, 2D masks often project onto artifacts and floaters within the Vanilla-NeRF model or erroneously pass through the object surface and project onto the background. Advancements in radiance field technology also lead to better segmentation results.}

In Tables~\ref{tab:spin-nerf} and \ref{tab:replica}, ``Single view'' refers to conducting mask inverse rendering exclusively for the 2D ground-truth mask of the reference view. This process is equivalent to mapping the 2D mask to the 3D space based on the corresponding depth information, without any subsequent learnable/updating step. We present these results to demonstrate the gain of the alternated process in our framework.
As shown in Table~\ref{tab:spin-nerf}, SA3D outperforms MVSeg~\cite{spinnerf} in most scenes. Taking SA3D-GS as an example, it achieves +9.6 mIoU on Truck and +17.1 mIoU on Lego. Besides, compared with the ``Single view'' model, a significant promotion is achieved, \textit{i.e.} +18.6 mIoU, which further proves the effectiveness of our method.

\begin{table*}[t!]
  \caption{Quantitative results on Replica (mIoU). `FC' denotes the feature cache.}
  \label{tab:replica}
  \centering
  \begin{tabular}{lcccccccc|cc}
  \toprule
    Scenes & office0 & office1 & office2 & office3 & office4 & room0 & room1 & room2 & mean & Time (s/object)\\
    \midrule
    Single view~\cite{sa3d} &68.7 & 56.5 & 68.4 & 62.2& 57.0 & 55.4& 53.8& 56.7 & 59.8 & -\\
    MVSeg~\cite{spinnerf} &  31.4  &  40.4    & 30.4 & 30.5 & 25.4 & 31.1 & 40.7 & 29.2 & 32.4 & 1430\\
    SA3D-TensoRF (w/o FC)~\cite{sa3d} & 84.4 & 77.0 & \textbf{88.9} & \textbf{84.4} & \textbf{82.6} & 77.6 & \textbf{79.8} & \textbf{89.2} & 83.0 & 67.6\\
    SA3D-TensoRF (w/ FC) & \textbf{86.2} & \textbf{78.2} & 87.2 & 83.6 & 81.5 & 79.1 & 77.5 & 88.1 & 82.7 & 30.2\\
    SA3D-GS (w/ FC) & 82.1 & 72.7 & 81.3 & 83.2 & 65.7 & \textbf{79.9} & 79.0 & 88.9 & 79.1 & 2.5\\
    \bottomrule
  \end{tabular}
\end{table*}

\subsubsection{The Replica Dataset}
\label{sec:replica}
We use the processed Replica data with 2D instance labels provided by Zhi \emph{et al.}~\cite{semantic-nerf} to evaluate the segmentation performance of SA3D. 
We retrieve all views containing each object and specify one reference view. 
With a similar setting of experiments on the SPIn-NeRF dataset, we use the ground-truth mask of the reference view and perform SA3D to conduct label propagation for evaluation. For each scene in Replica, around 20 objects are chosen for evaluation. 
As shown in Table~\ref{tab:replica}, the mean IoU (mIoU) is reported for all available objects in different scenes. We exclude the pixel-wise accuracy metric since an object only appears in a few views in the indoor scenes of Replica, where the pixel-wise accuracy is too high to serve as a reliable metric. {Besides, SA3D-NeRF is not evaluated on the Replica dataset, primarily due to its inferior reconstruction quality and notably slow inference speed: Assessing a single object with SA3D-NeRF on Replica takes over an hour, indicating vanilla-NeRF is unsuitable for segmenting scenes with a large number of objects.}

In complex indoor scenes of Replica, MVSeg's strategy based on cross-view 2D feature similarities proves to be ineffective, which generates numerous inaccurate 2D pseudo-labels, even using the Semantic-NeRF~\cite{semantic-nerf} for refinement. Consequently, the final 3D segmentation results of MVSeg even underperform those achieved by the ``Single view'' method. {SA3D-GS exhibits unexpectedly lower performance. We carefully check the results and find this is largely due to the sub-optimal geometry that 3D-GS learned from the Replica dataset, which manifests as numerous translucent artifacts that confuse the inverse-rendering process and, consequently, interfere with the self-prompting outcomes. This deficiency is likely due to the weak texture details provided by the Replica dataset, which result in poor point cloud initialization by Colmap~\cite{colmap1, colmap2}. During the training phase, incorrect 3D points become translucent floaters as 3D-GS tends to overfit certain views in the absence of multi-view consistency constraints, which are typically enforced by detailed textures. Improvements in 3D geometry in 3D-GS may overcome this challenge.}

\begin{figure*}
    \centering
    \includegraphics[width=\textwidth]{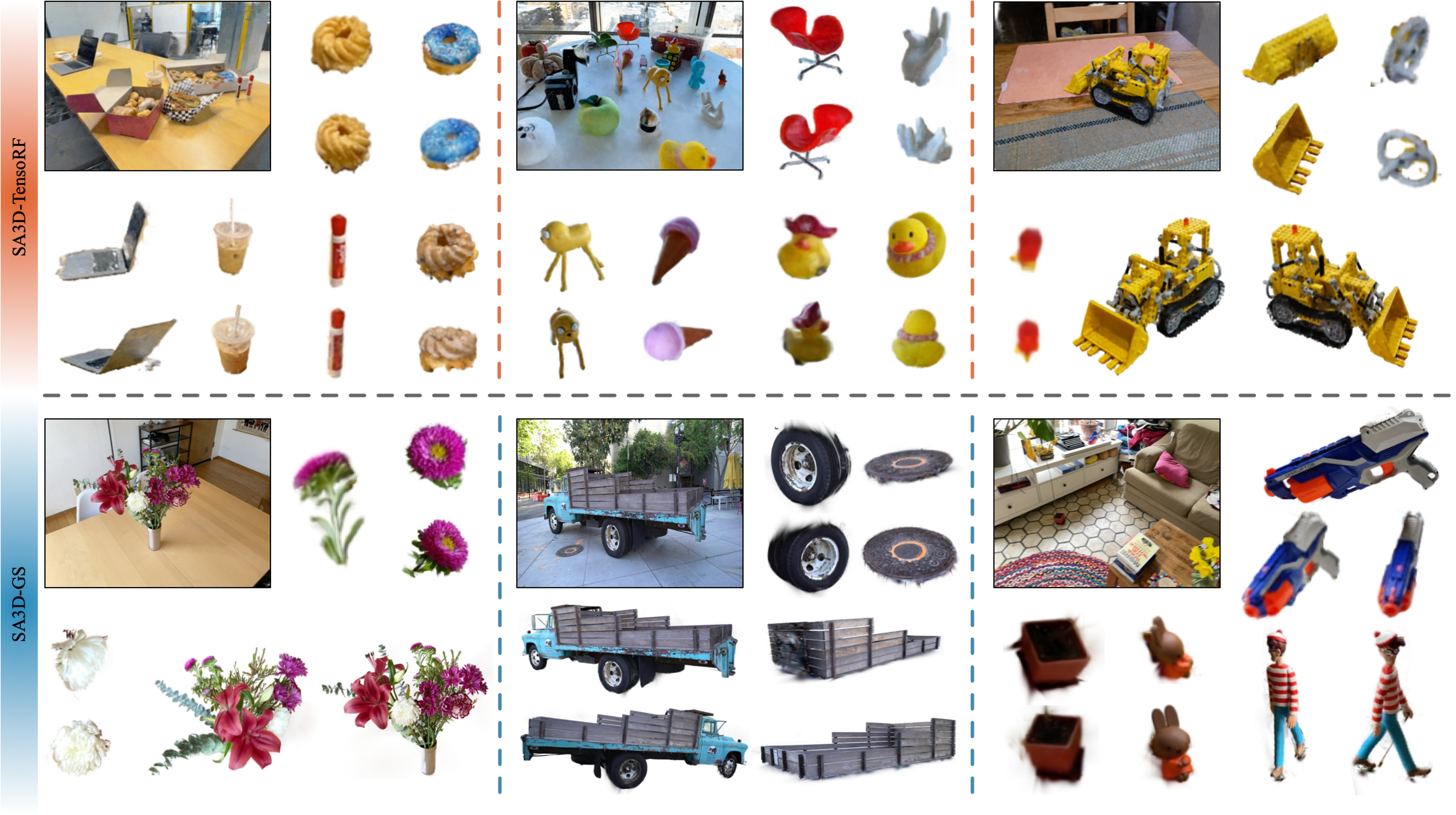}
    \caption{Some visualization results in different scenes (LERF-donuts~\cite{lerf}, LERF-figurines, 360-kitchen~\cite{mipnerf360}, LERF-bouquet, T\&T-truck~\cite{tanks} and LERF-nerfgun).}
    \label{fig:show}
\end{figure*}

\begin{figure}[t!]
    \centering
    \includegraphics[width=0.95\linewidth]{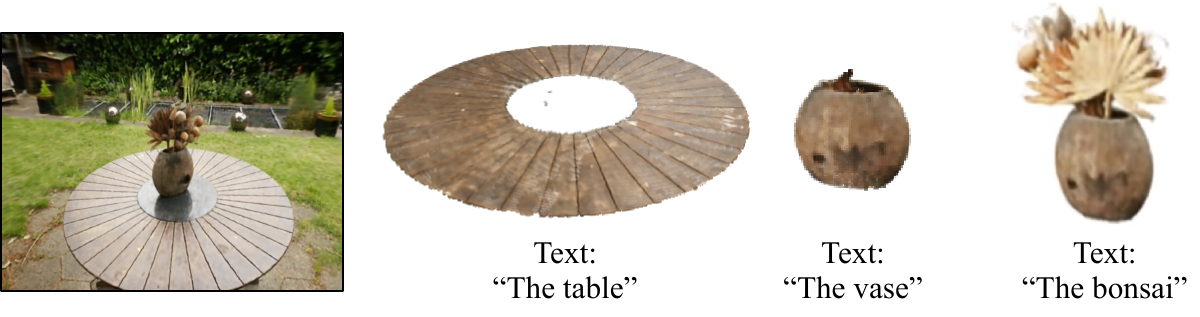}
    \caption{3D segmentation results of SA3D with the text prompts in 360-garden~\cite{mipnerf360}.}
    \label{fig:text_show}
\end{figure}

\begin{figure}[t!]
    \centering
    \includegraphics[width=0.95\linewidth]{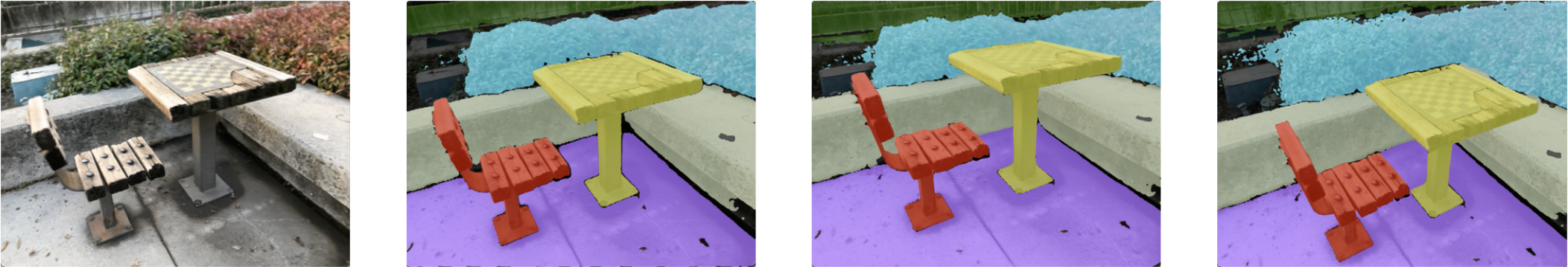}
    \caption{{Segmenting multiple objects simultaneously by extending the 3D mask to 4D in LLFF-chesstable~\cite{llff}.}}
    \label{fig:multi_obj}
\end{figure}

\subsection{Qualitative Results}
{We visualize the segmentation results of SA3D under three segmentation settings: object segmentation, part segmentation, and text-prompting segmentation. The first two are the core functions of SA3D. As shown in Figure~\ref{fig:show}, SA3D demonstrates its capability to segment diverse 3D objects across different scenes of different radiance fields, even when the objects are of small scales. Besides, SA3D can also handle challenging part segmentation. For instance, SA3D can segment distinct components of complex objects, such as the bucket, small wheel, and dome light of the Lego bulldozer. Furthermore, it exhibits a remarkable ability to accurately delineate individual elements from densely grouped objects, as evidenced by its precise segmentation of a single flower from within a bouquet.}
Figure~\ref{fig:text_show} demonstrates the potential of SA3D in combining with cross-modal models. Given a text phrase, the corresponding object can be accurately cut out. The text-prompting segmentation is built upon Grounding-DINO~\cite{grounding_dino}, a model capable of generating bounding boxes for objects based on text prompts. These bounding boxes serve as input prompts for SA3D in the segmentation process. {It is noteworthy that SA3D is capable of simultaneously segmenting multiple objects by augmenting the 3D mask with an additional dimension to account for distinct objects, as shown in Figure~\ref{fig:multi_obj}. 
} 

\begin{figure}
    \centering
    \includegraphics[width=0.95\linewidth]{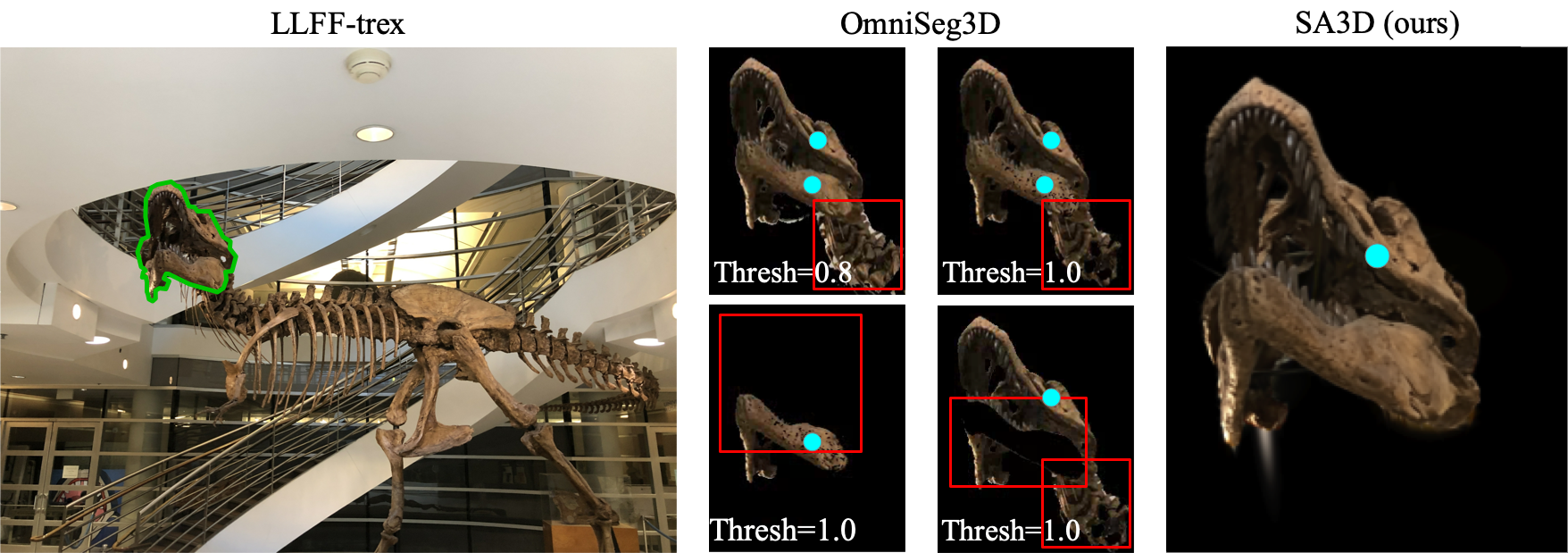}
    \caption{{Compared with OmniSeg3D (a feature-field-based radiance field segmentation approach), SA3D can successfully segment the T-rex skull, while OmniSeg fails. Target of interest is labeled with the \textcolor{greenframe}{green frame}, prompts are labeled as \textcolor{bluedots}{blue dots} and the failures of OmniSeg3D are labeled as \textcolor{red}{red boxes}.}}
    \label{fig:omnicompare}
\end{figure}

{In Figure~\ref{fig:omnicompare}, we demonstrate the superior alignment of SA3D with the specific requirements in comparison to feature-field-based approaches like OmniSeg3D~\cite{omniseg3d}. Despite experimenting with various prompts and adjusting the score threshold (a hyper-parameter in OmniSeg3D for filtering 3D points based on feature similarity), OmniSeg3D consistently fails to segment the T-rex skull accurately. As stated in Section~\ref{sec:seg_in_rfs}, this is because that the 2D masks used to train the feature field in these methods are generated automatically with SAM, which might not align with targets with unseen granularities. As a result, the trained feature field is certain to encounter failures when the required segmentation granularity differs from the granularities involved in training. Concretely, in the automated mask generation process, the T-rex skull is seldom segmented as a cohesive entity. Consequently, the feature fields trained using these automatically derived masks fail to segment the T-rex skull in its entirety. This limitation of feature-field-based methods has also been revealed by GARField~\cite{garfield}.}

\begin{table*}[htbp]
  \caption{Ablation on different numbers of views for 3D mask generation. Numbers in parentheses represent the view percentage of total training views. `FC' denotes the  feature cache.}
  \label{tab:ablation_num_views}
  \centering
  \begin{tabular}{c|c|cccc}
  \toprule
    \multicolumn{2}{c}{\multirow{2}{*}{Number of Views}} & 5 (10\%)  & 9 (20\%) & 21 (50\%) & 43 (100\%)\\
    \multicolumn{2}{c}{} & IoU (\%) / Time (s)&IoU (\%) / Time (s)&IoU (\%) / Time (s)&IoU (\%) / Time (s)\\
    \midrule
    \multirow{3}{*}{\makecell{Fortress \\ (forward facing)}} & SA3D-TensoRF (w/o FC)~\cite{sa3d}& 97.8 / 7.6 & 98.3 / 12.8 & 98.3 / 29.0 & 98.3 / 59.0\\
     & SA3D-TensoRF (w/ FC)& 98.0 / 6.0 &  98.4 / 7.4 & 98.5 / 12.8 & 98.4 / 20.3\\
      & SA3D-GS (w/ FC)& 97.2 / \textbf{1.6} & 98.1 / \textbf{2.2} & 98.1 / \textbf{2.5} & 98.1 / \textbf{3.7}\\
    \midrule
    \midrule
    \multicolumn{2}{c}{\multirow{2}{*}{Number of Views}} & 11 (10\%)  & 21 (20\%) & 51 (50\%) & 103 (100\%)\\
    \multicolumn{2}{c}{} & IoU (\%) / Time (s)&IoU (\%) / Time (s)&IoU (\%) / Time (s)&IoU (\%) / Time (s)\\
    \midrule
    \multirow{3}{*}{\makecell{Lego \\ (360$^\circ$)}} & SA3D-TensoRF (w/o FC)~\cite{sa3d}& 84.5 / 23.5& 84.8 / 43.5& 91.5 / 103.8& 92.2 / 204.9\\
     & SA3D-TensoRF (w/ FC)& 76.1 / 17.6 & 88.1 / 21.7 & 90.7 / 50.1 & 90.9 / 96.3 \\
      & SA3D-GS (w/ FC)& 84.4 / \textbf{1.9} & 88.2 / \textbf{2.3} & 88.8 / \textbf{3.2} & 92.0 / \textbf{4.6} \\
    \bottomrule
  \end{tabular}
\end{table*}

\subsection{Ablative Studies and Parameter Discussion}
In this section, we perform ablations to analyze the effectiveness of different modules and designs. Unless otherwise stated, SA3D refers to SA3D-TensoRF without the feature cache.
\label{sec:ablation}

\subsubsection{Number of Views} The process of mask inverse rendering and cross-view self-prompting is alternated across different views. By default, we utilize all available views in the training set $\mathcal{I}$. However, to expedite the 3D segmentation procedure, the number of views can be reduced. As shown in Table~\ref{tab:ablation_num_views}, We perform experiments on two representative scenes from the SPIn-NeRF~\cite{spinnerf} dataset to demonstrate this characteristic. The views are uniformly sampled from the sorted training set. In forward-facing scenes where the range of the camera poses is limited, satisfactory results can be achieved by selecting only a few views. {In 3D-GS, utilizing an NVIDIA RTX 3090 GPU, the 3D segmentation process across five views is finished within 2 seconds. Thanks to the feature cache, when views increase in a broader camera range, the segmentation results improve while the time cost increases negligibly. Increasing the view count from 11 to 103 brings only an additional 2.7 seconds in processing time.}

\subsubsection{Hyper-parameters} 
\begin{figure}[t!]
    \centering
    \includegraphics[width=\linewidth]{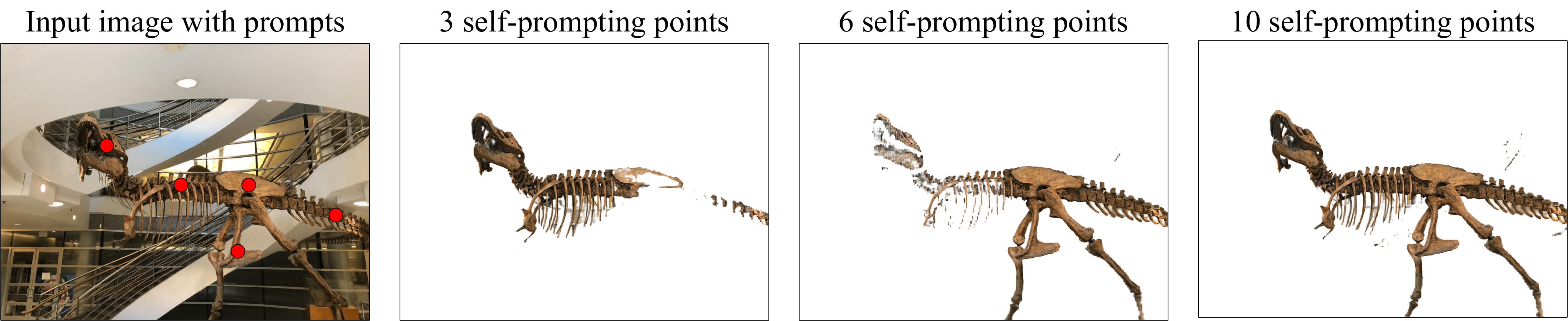}
    \caption{Results of different self-prompting points numbers $n_p$ on the LLFF-trex~\cite{llff} scene.
    }
    \label{fig:ablation_num_prompts}
\end{figure}

SA3D mainly involves three hyper-parameters: the IoU rejection threshold $\tau$, the loss balance coefficient $\lambda$ in Equation~\eqref{eq:loss_p2}, and the number of self-prompting points $n_p$. As shown in Table~\ref{tab:ablation_tau}, too small $\tau$ values lead to unstable SAM predictions, introducing noises to the 3D mask; too large $\tau$ values impede the 3D mask from getting substantial information.
Table~\ref{tab:ablation_lambda} indicates 
slightly introducing a negative term with the $\lambda$ factor can reduce noise for mask projection. However, a too-large negative term may make the mask completion process unstable and cause degraded performance. The selection of $n_p$ depends on the specific segmentation target, as SAM tends to produce over-segmented results that capture finer details of objects. As shown in Figure~\ref{fig:ablation_num_prompts}, for objects with a relatively large scale and complex structures, a
bigger $n_p$ produces better results. Empirically, setting $n_p$ to 3 can meet the requirements of most situations.

\begin{table}[htbp]
    \centering
    \caption{Ablation on different IoU-aware rejection threshold $\tau$ on the Replica office\_0.}
    \begin{tabular}{lccccc}
    \toprule
    $\tau$ & 0.3& 0.4& 0.5& 0.6 & 0.7 \\
    \midrule
    mIoU & 74.9& 79.7& 84.4& 81.3& 82.0 \\
    \bottomrule
  \end{tabular}
    \label{tab:ablation_tau}
\end{table}

\begin{table}[htbp]
    \centering
    \caption{Ablation on different negative term coefficient $\lambda$ on the Replica office\_0.}
    \begin{tabular}{lccccc}
    \toprule
    $\lambda$ & 0.05& 0.1& 0.15& 0.3 & 0.5 \\
    \midrule
    mIoU & 79.1 & 82.9& 84.4& 84.9& 83.3\\
    \bottomrule
  \end{tabular}
    \label{tab:ablation_lambda}
\end{table}

\subsubsection{Self-prompting Strategy} Without the 3D distance based confidence decay (Equation~\eqref{eq:decay}), our self-prompting strategy degrades to a simple 2D NMS (Non-Maximum Suppression), which selects a prompt point with the highest confidence score and then mask out a region around it. To show the efficacy of our design, we conduct experiments using the NVOS benchmark and present per-scene results for in-depth analysis.

Table~\ref{tab:ablation_confidence_decay} shows that a simple NMS self-prompting is enough for most cases. But for hard cases like `LLFF-trex' (a T-rex skeleton, as shown in Figure~\ref{fig:ablation_num_prompts}), where a large number of depth jumps, the confidence decay term contributes a lot. In such a situation, inaccurate masks bleed through gaps in the foreground onto the background. Suppose the self-prompting mechanism generates prompts on these inaccurate regions. In that case, SAM may produce plausible segmentation results that can cheat the IoU-rejection mechanism and finally, the segmentation results will involve unwanted background regions.

\begin{table}[htbp]
  \caption{Ablation on the confidence decay term of the self-prompting strategy.}
  \label{tab:ablation_confidence_decay}
  \centering
  \begin{tabular}{lcc|cc}
    \toprule
    \multirow{2}{*}{Scenes} &  \multicolumn{2}{c}{w/ Confidence Decay} & \multicolumn{2}{c}{w/o Confidence Decay}\\
    \cmidrule{2-5}
            & IoU (\%)     & Acc (\%) & IoU (\%)     & Acc (\%) \\
    \midrule
    Fern   & 82.9 & 94.4 & 82.9  & 94.4\\
    Flower & 94.6 & 98.7 & 94.6 & 99.7\\
    Fortress     & 98.3 & 99.7 & 98.4 & 99.7\\
    Horns-center & 96.2 & 99.3 & 96.2 & 99.3 \\
    Horns-left    & 90.2 & 99.4 & 88.8 & 99.3  \\
    Leaves & 93.2& 99.6 & 93.2& 99.6  \\
    Orchids     & 85.5 & 97.3 & 85.4 & 97.3  \\
    Trex & 82.0 & 97.4 & 64.0 & 93.3\\
    \midrule
    mean & 90.3& 98.2& 87.9& 97.7\\
    \bottomrule
  \end{tabular}
\end{table}

\subsubsection{2D Segmentation Models} In addition to SAM, we also incorporate four other prompt-based 2D segmentation models~\cite{seem, sofiiuk2022reviving, liu2023simpleclick, chen2022focalclick} into our framework to demonstrate the generalization ability of SA3D. The evaluation results on the NVOS dataset are shown in Table~\ref{tab:ablation_seg_model}. With different 2D segmentation models, SA3D performs stably. Though the performance of SA3D with SEEM~\cite{seem} is somewhat lower than with SAM, it can support more kinds of input prompts like reference images. These results indicate the potential of SA3D for integration with diverse segmentation models to support various 3D segmentation tasks.

\begin{table*}[htbp]
  \caption{Ablation on different 2D segmentation models.}
  \label{tab:ablation_seg_model}
  \centering
  \begin{tabular}{cccccccccc}
    \toprule
    \multicolumn{2}{c}{SAM~\cite{sam}} & \multicolumn{2}{c}{SEEM~\cite{seem}} & \multicolumn{2}{c}{SimpleClick~\cite{liu2023simpleclick}} & \multicolumn{2}{c}{RITM~\cite{sofiiuk2022reviving}} & \multicolumn{2}{c}{FocalClick~\cite{chen2022focalclick}}\\
    \cmidrule{1-10}
     mIoU (\%)     & mAcc (\%) & mIoU (\%)     & mAcc (\%) & mIoU (\%)     & mAcc (\%) &mIoU (\%)     & mAcc (\%)&mIoU (\%)     & mAcc (\%) \\
     \textbf{90.3} & \textbf{98.2} & 86.0 & 97.0 & 87.7 & 97.8 & 81.2 & 96.3 & 88.9 & 98.1 \\
    \bottomrule
  \end{tabular}
\end{table*}

\subsubsection{Time Cost Analysis with Different Representations}
{In Section~\ref{sec:quantitative}, we show that SA3D performs stably when integrated with different radiance fields. In this section, we focus on analyzing the time consumption associated with these different radiance fields. Table~\ref{tab:concrete_time} illustrates that, across both forward-facing and 360$^{\circ}$ scenes, SA3D with 3D-GS achieves optimal efficiency. With the feature cache disabled, the primary bottleneck for SA3D-GS shifts to the self-prompting phase, particularly due to the time-intensive SAM encoder forward pass. For vanilla-NeRF and TensoRF, time is mainly spent on mask rendering and mask inverse rendering. The segmentation speed of SA3D-GS, exceeding 10 frames per second, approaches real-time performance. 
Though the effectiveness of the feature cache, it is important to note that the feature cache is designed to expedite the segmentation of multiple objects within the same scene. When focusing on a single object, the total time for feature extraction and segmentation aligns closely with using SA3D without the feature cache.}

\begin{table*}[htbp]
    \centering
    \caption{Analysis of time consumption with different kinds of radiance fields. `FC' denotes the feature cache. The `Self-prompting' phase involves generating prompts, querying SAM (or SAM decoder) for segmentation, and loss calculation.}
    \begin{tabular}{ccccccc}
    \toprule
       \multirow{2}{*}{Scene}  &  \multirow{2}{*}{Radiance Field} & \multirow{2}{*}{Resolution} & \multirow{2}{*}{Speed (fps) $\uparrow$} & \multicolumn{3}{c}{Time per Step (ms)$\downarrow$}\\
       \cmidrule{5-7}
       & & & & Mask Rendering & Self-prompting & Inverse Rendering\\
       \midrule
       \multirow{4}{*}{Horns~\cite{llff}} &  Vanilla-NeRF (w/ FC)& $378\times504$ & $ 0.2$ & $ 4211.8$ & $ 12.5$& $ 249.9$\\
        & TensoRF (w/ FC)& $756\times1008$ & $ 2.9$ & $ 263.7$ & $ 18.1$& $ 65.1$\\
        & 3D-GS (w/ FC)& $1200\times1600$  & $ 17.7$& $ 8.3$ & $ 40.7$ & $ 6.4$\\
        & 3D-GS (w/o FC)& $1200\times1600$ & $ 1.9$ & $ 8.3$ & $ 508.1$ & $ 6.4$\\
       \midrule
       \multirow{4}{*}{Pinecone~\cite{tanks}} &  Vanilla-NeRF (w/ FC) & $363\times484$ & $ 0.2$ & $ 3909.4$ & $ 12.0$& $ 229.0$\\
        & TensoRF (w/ FC) &  $363\times484$ & $4.3$ & $195.8$ & $17.7$ & $15.9$\\
        & 3D-GS (w/ FC) & $1200 \times 1600$  & $ 13.2$& $ 10.7$ & $ 48.4$& $ 15.7$\\
        & 3D-GS (w/o FC) & $1200 \times 1600$  & $ 1.9 $& $ 10.7 $ & $ 515.8$ & $ 15.7$\\
        \bottomrule
    \end{tabular}
    \label{tab:concrete_time}
\end{table*}

\subsubsection{Resolution of Mask Grids}
{We explore the impact of the resolution of 3D mask grids $\mathbf{V}$ on segmentation performance. As indicated in Table~\ref{tab:ablation_3d_mask}, our evaluation spans the spectrum from varying resolutions of mask grids to employing a Multi-Layer Perceptron (MLP) for representing the 3D mask, the latter of which theoretically offers infinite resolution. Our findings reveal that the resolution of mask grids has a negligible influence on segmentation outcomes. This minimal impact is attributed to the inherent characteristics of segmentation masks: if a point is located within a mask, it is very likely to be considered part of the corresponding segment. Thus, adopting low resolution grids with simple trilinear interpolation is enough to obtain a competitive segmentation result. Conversely, adopting an MLP for the 3D mask representation significantly hampers performance. This decline is linked to the challenges in achieving stable optimization with implicit representations. SA3D's effectiveness relies on robust optimization, as its cross-view self-prompting mechanism depends on the rapid convergence of mask representation. The process involves using the 3D mask to immediately render a 2D mask in a novel view after a single iteration of mask inverse rendering. The MLP's behavior is unstable due to the intertwined nature of its parameters, complicating control over the outcome.
This analysis indicates that for NeRFs, a set of explicit mask grids with proper resolution is the most suitable 3D mask representation.}

\begin{table*}[htbp]
  \caption{Ablation on the resolution of 3D mask grids.}
  \label{tab:ablation_3d_mask}
  \centering
  \begin{tabular}{cccccccccc}
    \toprule
    \multicolumn{2}{c}{$320^3$ Grids (default)} & \multicolumn{2}{c}{$160^3$ Grids} & \multicolumn{2}{c}{$80^3$ Grids} & \multicolumn{2}{c}{$40^3$ Grids} & \multicolumn{2}{c}{MLP}\\
    \cmidrule{1-10}
     mIoU (\%)     & mAcc (\%) & mIoU (\%)     & mAcc (\%) &mIoU (\%)     & mAcc (\%)&mIoU (\%)     & mAcc (\%) &mIoU (\%)     & mAcc (\%) \\
     $\mathbf{90.3}$ & $\mathbf{98.2}$ & 90.1 & 98.2 & 89.2 & 98.1 & 88.0 & 97.8 & 80.9 & 96.4 \\
    \bottomrule
  \end{tabular}
\end{table*}

\subsubsection{Ambiguous Gaussians Removal for SA3D-GS}

\begin{figure}
    \centering
    \includegraphics[width=\linewidth]{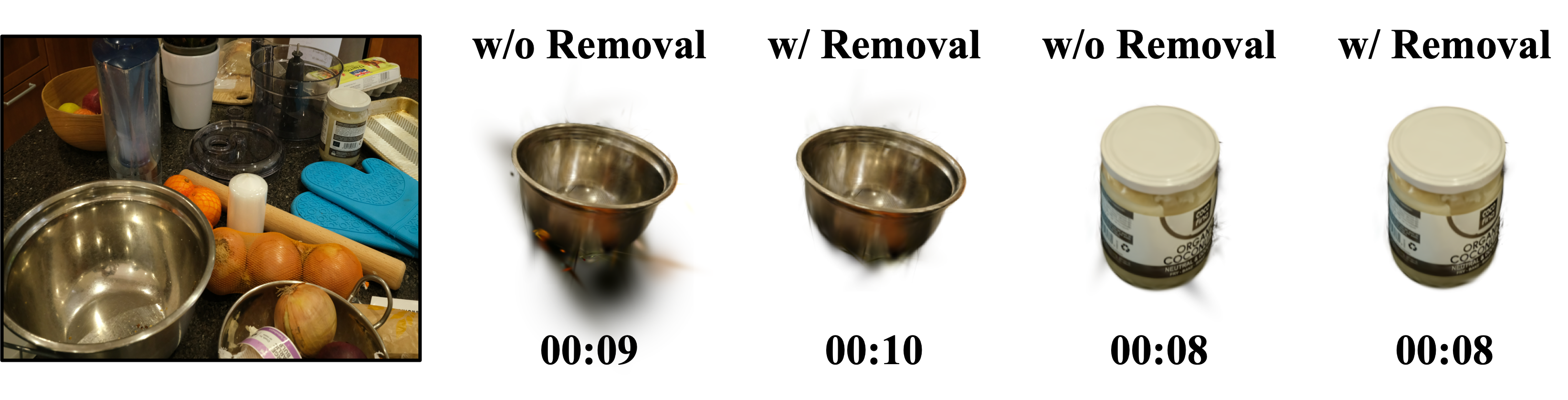}
    \caption{The effect of the ambiguous Gaussians removal strategy for SA3D-GS. Experiments are conducted on 360-counter~\cite{mipnerf360}. Corresponding segmentation time cost is labelled below each segmentation result. This strategy can eliminate ambiguous Gaussians at the border of the segmented target within one second.}
    \label{fig:ablation_artifacts_removal}
\end{figure}

{In Section~\ref{sec:ambi} we introduce an ambiguous Gaussians removal strategy for SA3D-GS. We provide more qualitative results to illustrate and demonstrate its effectiveness intuitively. Results are shown in Figure~\ref{fig:ablation_artifacts_removal}. Our experiments point out that there are more ambiguous Gaussians at the border of two objects that share similar color. For example, due to the reflective effect of the metal basin, it appears a similar color to the table below. After removing the background the segmented basin has more artifacts than the segmented jar. This issue roots in the optimization goal of 3D-GS. The training objective of 3D-GS is to fit the RGB color of multi-view images. Thus, for adjacent objects sharing similar color, 3D-GS tends to use one or few Gaussians to model their border.}

{This supplementary phase of ambiguous Gaussians removal bypasses the requirements for the SAM decoder, cross-view self-prompting, and IoU computations. It can be finished within one second and thus bring negligible additional time to the overall pipeline. However, this simplicity has its limitations. Some ambiguous Gaussians contribute significantly to rendering the segmentation target, with the occluded parts being relatively small. As a result, during the removal phase, the received gradient from the negative refinement term is still smaller than the positive term. Consequently, these Gaussians cannot be eliminated naively. Nevertheless, our simple initial exploration provides a good trade-off between effectiveness and computational efficiency.}

\begin{figure}[t!]
    \centering
    \includegraphics[width=\linewidth]{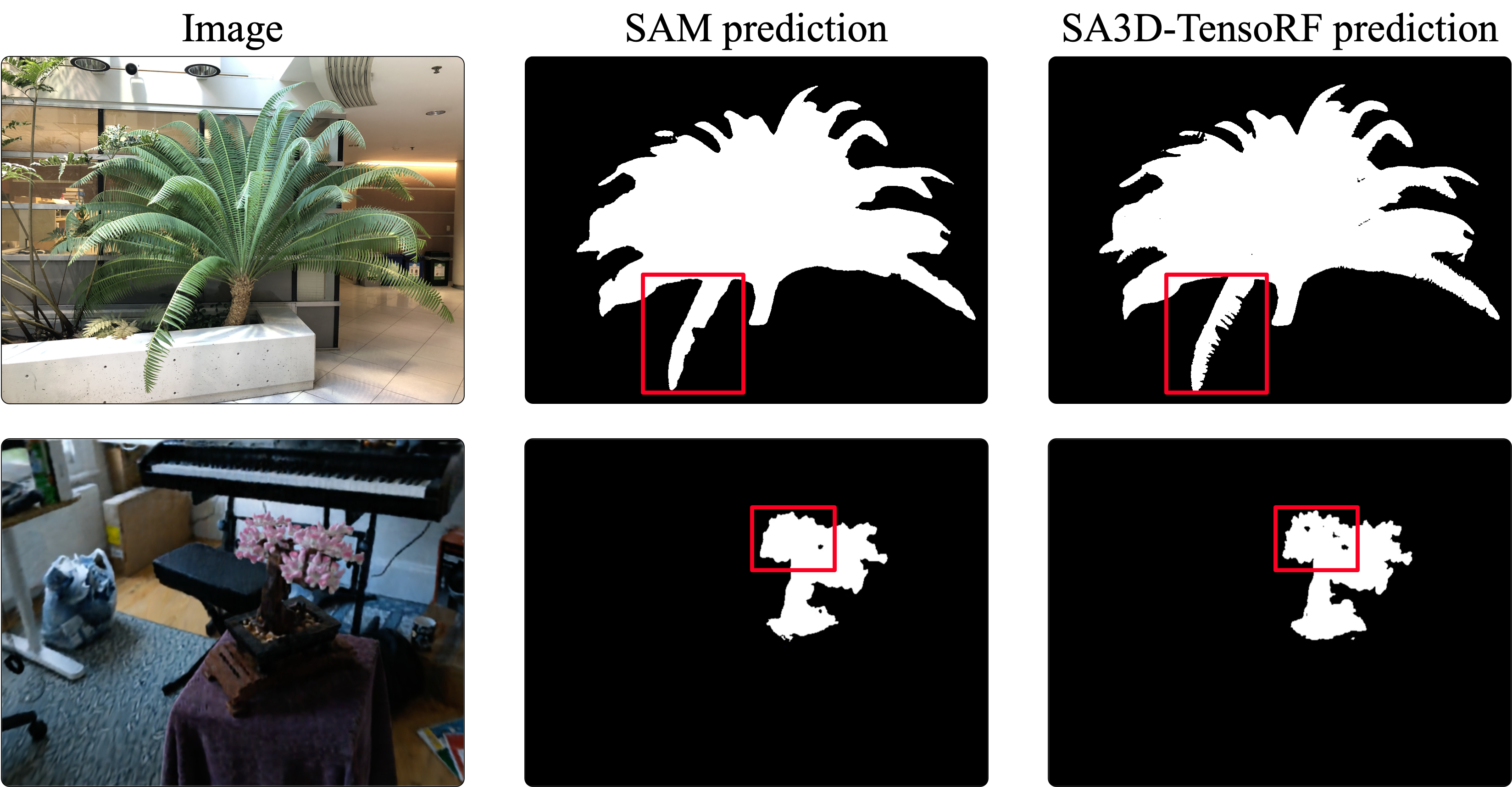}
    \caption{The 2D segmentation result by SAM and the 3D segmentation result by SA3D of the LLFF-fern~\cite{llff} scene and 360-bonsai~\cite{mipnerf360} scene. SA3D produces more details that are missing in the SAM segmentation result.}
    \label{fig:nerf_helps_sam}
\end{figure}

\begin{figure}[t!]
    \centering
    \includegraphics[width=\linewidth]{figures/SA3D-pami-artifacts-analysis.pdf}
    \caption{Comparison of the reconstruction quality of TensoRF~\cite{tensorf} and 3D-GS~\cite{3dgs} in Replica-office4~\cite{replica}. After 2D masks are projected onto the translucent artifacts, it will be rendered wrongly in another view, explaining the poor performance of SA3D-GS shown in Table~\ref{tab:replica}.}
    \label{fig:bad3dgs}
\end{figure}

\section{Discussion}
\label{sec:dis}
We provide more discussions to better understand the working mechanism of SA3D.

{First, \textbf{radiance fields can help to improve the segmentation quality of SAM}. In Figure~\ref{fig:nerf_helps_sam}, we show that SA3D can eliminate segmentation errors of SAM and effectively capture details such as holes and edges. 2D segmentation models often face challenges in accurately capturing object details such as small holes and gaps due to limitations in resolution. Even though SAM exhibits fine-grained segmentation capabilities, this issue still persists. The ability to improve SAM's performance stems from the fine-grained depth estimation (or the geometry information) provided by radiance fields. The utilization of such information to aid segmentation has been a long-standing problem~\cite{zhang2018joint, semantic-nerf, lerf}. Specific to SA3D, the geometry information is utilized through the incorporation of a negative refinement term in the projection loss (Equation~\eqref{eq:loss_p2}). When SAM overlooks small holes and gaps, the mask passes through these regions and gets projected onto the background behind the object. However, with a viewpoint switched, these inaccurately-segmented regions shift from being behind the target object to the side. In these new views, SAM's foreground prediction no longer includes these regions. Consequently, the mask confidence score for these regions is effectively suppressed by the negative refinement term.}

{However, from the other side, this phenomenon also implies that \textbf{wrongly learned geometry will confuse the segmentation model and lead to worse performance}. As shown in Table~\ref{tab:replica}, the performance of SA3D-GS is slightly below the expectation. In Figure~\ref{fig:bad3dgs}, we show that this degradation is mainly caused by the artifacts learned by 3D-GS. As a synthesized dataset, the rendered images in Replica~\cite{replica} lack detailed textures that guide the reconstruction of both Colmap~\cite{colmap1,colmap2} (the technique utilized by 3D-GS to obtain initial point clouds from the multi-view images) and 3D-GS. This indulges 3D-GS in overfitting particular views and generating numerous translucent artifacts in the space. During mask inverse rendering, the 2D masks are projected onto these artifacts and rendered wrongly in other views, leading to bad performance.}

\section{Conclusion}
\label{sec:conclusion}
In this paper, we propose SA3D, a framework that generalizes 2D SAM to segment 3D objects with radiance fields as the structural prior. Based on any trained radiance field and a set of prompts in a single view, SA3D performs an iterative procedure that involves rendering novel 2D views, self-prompting SAM for 2D segmentation, and projecting the segmentation back into the 3D space to refine a 3D mask. Extensive experiments are conducted for analysis, offering valuable perspectives on the development of segmentation foundation models and radiance fields. Our research sheds light on a resource-efficient methodology that lifts vision foundation models from 2D to 3D. Our future work may include improving learned 3D geometry of radiance fields with the help of segmentation models and supporting automatic panoptic segmentation.

\section*{Acknowledgments}
This work was supported by NSFC 62322604, NSFC 62176159, Natural Science Foundation of Shanghai 21ZR1432200, and Shanghai Municipal Science and Technology Major Project 2021SHZDZX0102. We thank Dr. Weichao Qiu for his insightful suggestions for this work.

\bibliographystyle{IEEEtran}
\bibliography{ref}

\end{document}